%% file: main.tex
\documentclass[a4paper]{article}

\usepackage[english]{babel}
\usepackage[utf8x]{inputenc}
\usepackage[T1]{fontenc}

\usepackage{amsmath}
\usepackage{amssymb}
\usepackage{comment}
\usepackage{graphicx}
\usepackage{cancel}
\usepackage{amsthm}
\usepackage{dsfont}
\usepackage{xcolor}
\usepackage{ stmaryrd }
\usepackage{enumitem} 

\usepackage{setspace}
\usepackage{enumitem}
\usepackage{wrapfig}
\usepackage{cite}

\usepackage{graphicx} 
\usepackage{subfigure} 
\usepackage[a4paper,top=3cm,bottom=2cm,left=3cm,right=3cm,marginparwidth=1.75cm]{geometry}

\usepackage[algo2e,inoutnumbered,linesnumbered,algoruled,vlined]{algorithm2e}
\usepackage[colorlinks=true, allcolors=blue]{hyperref}

\input{defs}

\graphicspath{{./},{./figures/}}

\title{A Tail-Index Analysis of Stochastic Gradient Noise in Deep Neural Networks}
\author{Umut \c Sim\c sekli\thanks{LTCI, T\'{e}l\'{e}com ParisTech, Universit\'{e} Paris-Saclay, 75013, Paris, France.},
\c Levent Sagun\thanks{Institute of Physics, \'{E}cole Polytechnique F\'ed\'erale de Lausanne,
1015 Lausanne, Switzerland.},
Mert G\"urb\"uzbalaban \thanks{Department of Management Science and Information Systems, Rutgers Business School, NJ 08854, USA.}}
\date{}
\begin{document} 
 
\maketitle 
\input{abstract}

\input{intro}

\input{levy_sde}

\input{experiments}

\input{conclusion}

\section*{Acknowledgments}

This work is partly supported by the French National Research Agency (ANR) as a part of the FBIMATRIX (ANR-16-CE23-0014) project. Mert G\"urb\"uzbalaban acknowledges support from the grants NSF DMS-1723085 and NSF CCF-1814888.

\bibliography{sgd_tail,work}
\bibliographystyle{plain}

\end{document}


\onecolumn
\icmltitle{A Tail Index Analysis of Stochastic Gradient Noise in Deep Neural Networks \\ {\large SUPPLEMENTARY DOCUMENT}}



 
\begin{icmlauthorlist}
\icmlauthor{Umut \c Sim\c sekli}{tpt}
\icmlauthor{\c Ca\u{g}atay Y{\i}ld{\i}z}{aalto}
\icmlauthor{Thanh Huy Nguyen}{tpt}
\icmlauthor{Ga\"{e}l Richard}{tpt} 
\icmlauthor{A. Taylan Cemgil}{boun}
\end{icmlauthorlist}

\icmlaffiliation{aalto}{Department of Computer Science, Aalto University, Espoo, 02150, Finland}
\icmlaffiliation{boun}{Department of Computer Engineering, Bo\u{g}azi\c ci University, 34342, Bebek, Istanbul, Turkey}
\icmlaffiliation{tpt}{LTCI, T\'{e}l\'{e}com ParisTech, Universit\'{e} Paris-Saclay, 75013, Paris, France}

\icmlcorrespondingauthor{Umut \c Sim\c sekli}{umut.simsekli@telecom-paristech.fr}


\icmlkeywords{mcmc}

\vskip 0.3in



\printAffiliationsAndNotice{}  

\input{alpha_estim}

\input{width_flat}

\input{discretization}


\bibliography{sgd_tail,work}
\bibliographystyle{icml2019}

%% file: defs.tex
\newcommand{\sil}[1]{}

\newcommand{\wb}{\mathbf{w}}
\newcommand{\rset}{\mathbb{R}}
\newcommand{\E}{\mathbb{E}}

\newcommand{\rmd}{\mathrm{d}}
\newcommand{\Bm}{\mathrm{B}}
\newcommand{\Lm}{\mathrm{L}^\alpha}
\newcommand{\Id}{\mathbf{I}}
\newcommand{\beq}{\begin{eqnarray}}
\newcommand{\eeq}{\end{eqnarray}}
\newcommand{\beqs}{\begin{eqnarray*}}
\newcommand{\eeqs}{\end{eqnarray*}}

\newcommand{\sas}{{\cal S}\alpha{\cal S}}

\newtheorem{thm}{Theorem}

\DeclareMathOperator*{\argmin}{arg\min}

%% file: abstract.tex

\begin{abstract} 
The gradient noise (GN) in the stochastic gradient descent (SGD) algorithm is often considered to be Gaussian in the large data regime by assuming that the \emph{classical} central limit theorem (CLT) kicks in. 
This assumption is often made for mathematical convenience, since it enables SGD to be analyzed as a stochastic differential equation (SDE) driven by a Brownian motion.
We argue that the Gaussianity assumption might fail to hold in deep learning settings and hence render the Brownian motion-based analyses inappropriate. Inspired by non-Gaussian natural phenomena, we consider the GN in a more general context and invoke the \emph{generalized} CLT (GCLT), which suggests that the GN converges to a \emph{heavy-tailed} $\alpha$-stable random variable.
Accordingly, we propose to analyze SGD as an SDE driven by a L\'{e}vy motion. Such SDEs can incur `jumps', which force the SDE \emph{transition} from narrow minima to wider minima, as proven by existing metastability theory. 
To validate the $\alpha$-stable assumption, we conduct extensive experiments on common deep learning architectures and show that in all settings, the GN is highly non-Gaussian and admits heavy-tails. We further investigate the tail behavior in varying network architectures and sizes, loss functions, and datasets. Our results open up a different perspective and shed more light on the belief that SGD prefers wide minima.
%
%
%
%
%
%
%
\end{abstract} 

%

%% file: intro.tex

\section{Introduction}
\label{sec:intro}

\textbf{Context and motivation: }
Deep neural networks have revolutionized machine learning and have ubiquitous use in many application domains \cite{hinton-nature,Krizhevsky12,Hinton12}. 
In full generality, many key tasks in deep learning reduces to solving the following optimization problem:
\begin{align}
\wb^\star = \argmin_{\wb \in \rset^p} \Bigl\{ f(\wb) \triangleq \frac1{n} \sum\nolimits_{i=1}^n f^{(i)}(\wb) \Bigr\}
\end{align}
where $\wb\in\rset^p$ denotes the weights of the neural network, $f:\rset^p\to\rset$ denotes the loss function that is typically non-convex in $\wb$, each $f^{(i)}$ denotes the (instantaneous) loss function that is contributed by the \emph{data point} $i \in \{1, \dots, n\}$, and $n$ denotes the total number of data points.
Stochastic gradient descent (SGD) is one the most popular approaches for attacking this problem in practice and is based on the following iterative updates:
\begin{align}
 \wb_{k+1} = \wb_{k} - \eta \nabla \tilde{f}_{k}(\wb_k) \label{eqn:sgd_main}
\end{align} 
where $k \in \{1, \dots, K\}$ denotes the iteration number and $\nabla \tilde{f}_{k}$ denotes the stochastic gradient at iteration $k$, that is defined as follows:
\begin{align}
\label{eqn:stoch_grad}
\nabla \tilde{f}_{k} (\wb) \triangleq \nabla \tilde{f}_{\Omega_k} (\wb) \triangleq \frac1{b} \sum\nolimits_{i \in \Omega_k}  \nabla f^{(i)}(\wb).
\end{align} 
Here, $\Omega_k \subset \{1,\dots,n\}$ is a random subset that is drawn with or without replacement at iteration $k$, and $b = |\Omega_k|$ denotes the number of elements in $\Omega_k$. 

SGD is widely used in deep learning with a great success in its computational efficiency \cite{bottou2010large,bottou2008tradeoffs}. Beyond efficiency, understanding how SGD performs better than its full batch counterpart in terms of test accuracy remains a major challenge. Even though SGD seems to find zero loss solutions on the training landscape (at least in certain regimes \cite{Zhang16, sagun2014explorations, keskar2016large, Geiger18}), it appears that the algorithm finds solutions with different properties depending on how it is tuned \cite{sutskever2013importance, keskar2016large, jastrzkebski2017three, hoffer2017train, masters2018revisiting, smith2017don}. Despite the fact that the impact of SGD on generalization has been studied \cite{advani2017high, wu2018sgd, neyshabur2017exploring}, a satisfactory theory that can explain its success in a way that encompasses such peculiar empirical properties is still lacking. 

A popular approach for analyzing SGD is based on considering SGD as a discretization of a continuous-time process \cite{mandt2016variational,jastrzkebski2017three,pmlr-v70-li17f,hu2017diffusion,zhu2018anisotropic,chaudhari2018stochastic}. This approach mainly requires the following assumption\footnote{We note that more sophisticated assumptions than \eqref{eqn:noise_gauss} have been made in terms of the covariance matrix of the Gaussian distribution (e.g.\ state dependent, anisotropic). However, in all these cases, the resulting distribution is still a Gaussian, therefore the same criticism holds.} on the stochastic gradient noise $U_k(\wb) \triangleq \nabla \tilde{f}_{k} (\wb) - \nabla f(\wb)$:
\begin{align}
U_k(\wb) \sim {\cal N}(\mathbf{0}, \sigma^2 \Id), \label{eqn:noise_gauss}
\end{align}
where ${\cal N}$ denotes the multivariate (Gaussian) normal distribution and $\Id$ denotes the identity matrix of appropriate size. The rationale behind this assumption is that, if the size of the minibatch $b$ is large enough, then we can invoke the Central Limit Theorem (CLT) and assume that the distribution of $U_k$ is approximately Gaussian. Then, under this assumption, \eqref{eqn:sgd_main} can be written as follows:
\begin{align}
\wb_{k+1} = \wb_{k} - \eta \nabla f(\wb_k) + \sqrt{\eta} \sqrt{\eta \sigma^2} Z_k, \label{eqn:sgd_gauss}
\end{align}
where $Z_k$ denotes a standard normal random variable in $\rset^p$. If we further assume that the step-size $\eta$ is small enough, then the continuous-time analogue of the discrete-time process \eqref{eqn:sgd_gauss} is the following stochastic differential equation (SDE):\footnote{
In a recent work with a similar critic taken on the recent theories on the SGD dynamics, some theoretical concerns have been also raised about the SDE approximation of SGD \cite{yaida2018fluctuationdissipation}. We believe that the SDE representation is sufficiently accurate for small step-sizes and a good, if not the best, proxy for understanding the behavior of SGD.
%
}
\begin{align}
\rmd \wb_t = - \nabla f(\wb_t) \rmd t + \sqrt{\eta \sigma^2} \rmd \Bm_t , \label{eqn:sgd_langevin}
\end{align}
where $\Bm_t$ denotes the standard Brownian motion. This SDE is a variant of the well-known \emph{Langevin diffusion} and under mild regularity assumptions on $f$, one can show that the Markov process $(\wb_t)_{t\geq 0}$ is ergodic with its unique invariant measure, whose density is proportional to $\exp(-f(x)/(\eta \sigma^2))$ for any $\eta>0$. \cite{Roberts03}. From this perspective, the SGD recursion in \eqref{eqn:sgd_gauss} can be seen as a first-order Euler-Maruyama discretization of the Langevin dynamics (see also \cite{pmlr-v70-li17f,jastrzkebski2017three,hu2017diffusion}), which is often referred to as the Unadjusted Langevin Algorithm (ULA) \cite{Roberts03,lamberton2003recursive,durmus2015non}. 

Based on this observation, \cite{jastrzkebski2017three} focused on the relation between this invariant measure and the algorithm parameters, namely the step-size $\eta$ and mini-batch size, as a function of $\sigma^2$. They concluded that the ratio of learning rate divided by the batch size is the control parameter that determines the width of the minima found by SGD. Furthermore, they revisit the famous wide minima folklore \cite{hochreiter1997flat}: Among the minima found by SGD, the wider it is, the better it performs on the test set. However, there are several fundamental issues with this approach, which we will explain below. 

\begin{figure}[t]
    \centering
    \subfigure[Real]{\includegraphics[width=0.32\columnwidth]{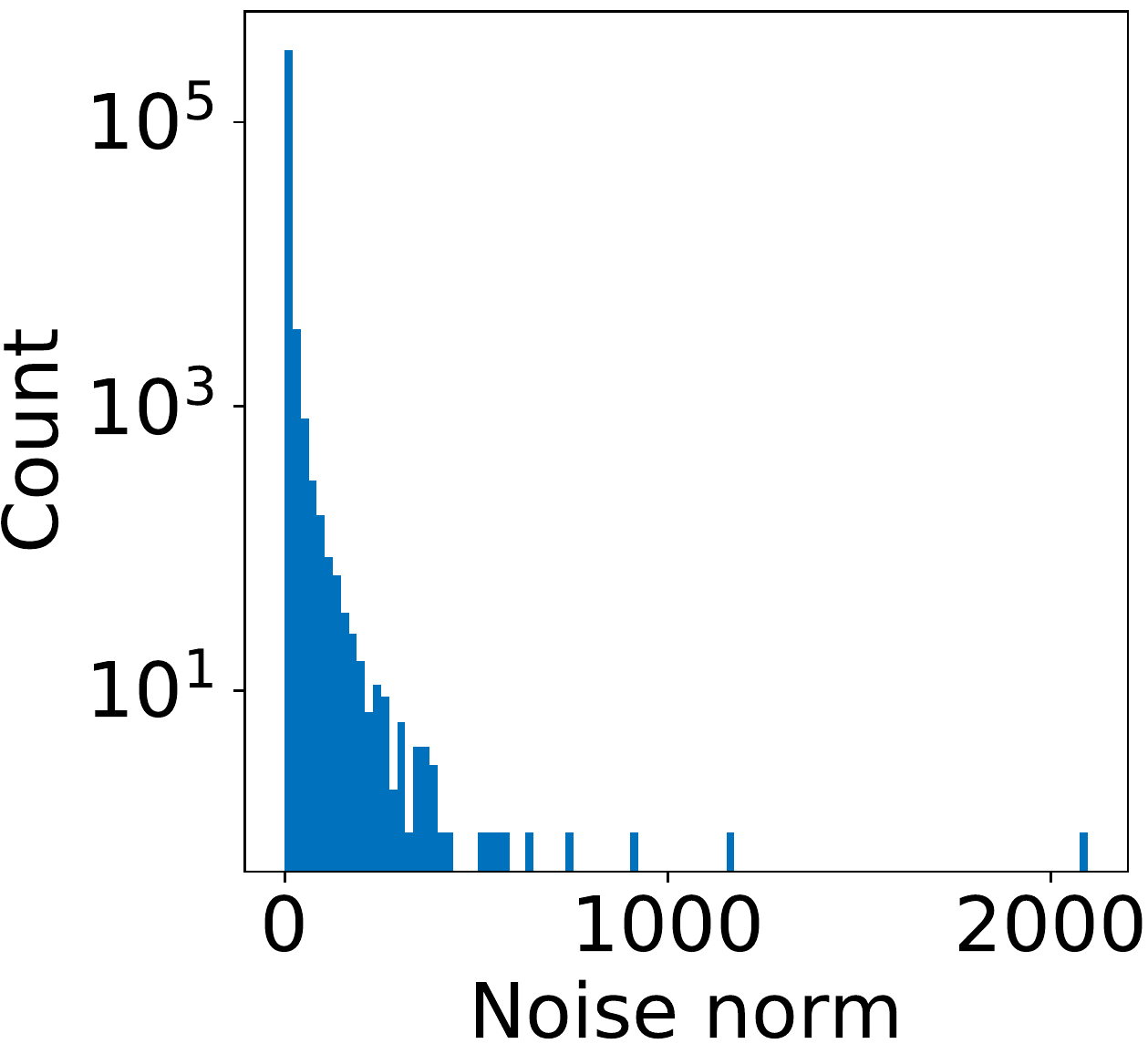}}
    \subfigure[Gaussian]{\includegraphics[width=0.32\columnwidth]{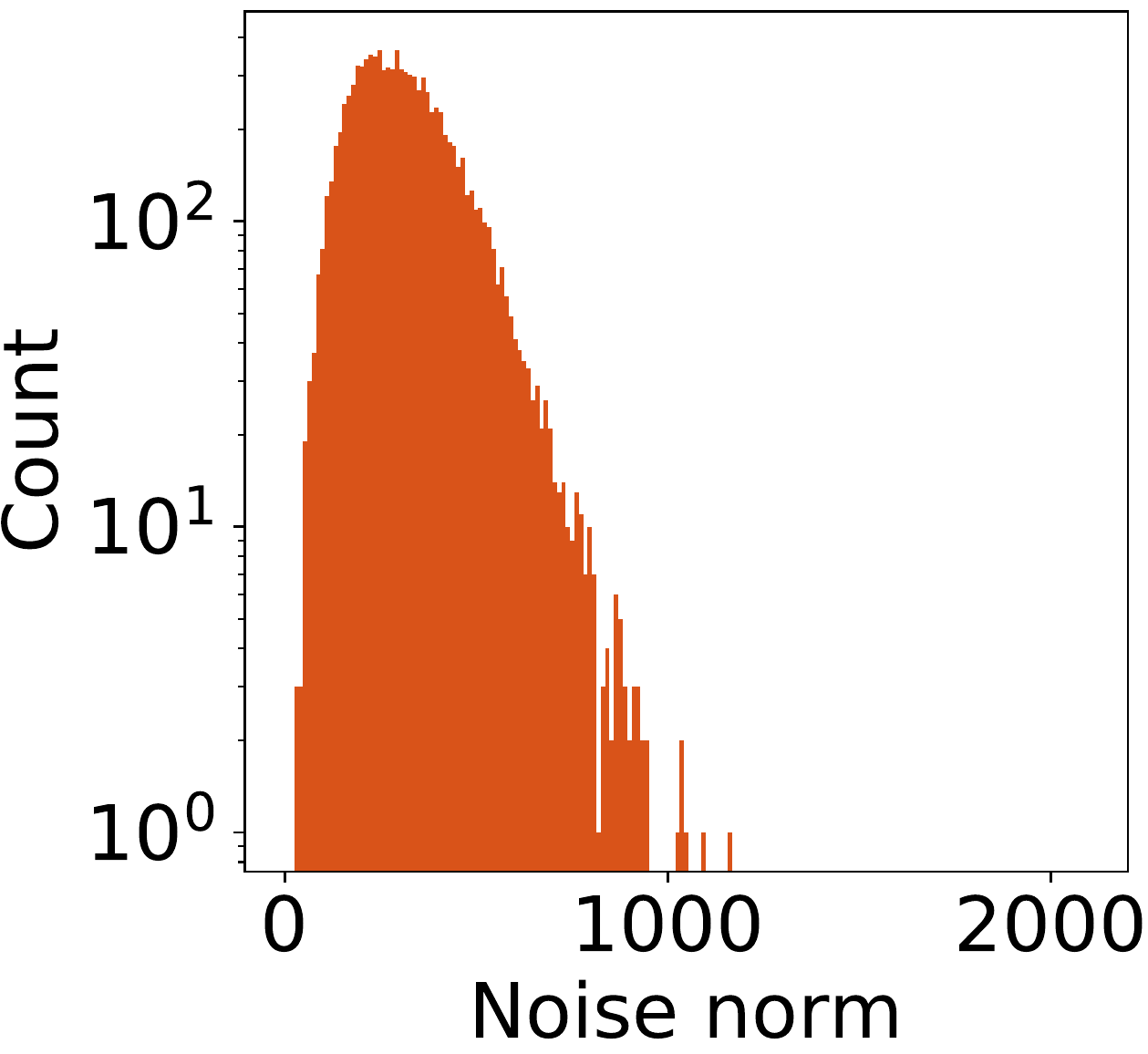}}
    \subfigure[$\alpha$-stable]{\includegraphics[width=0.32\columnwidth]{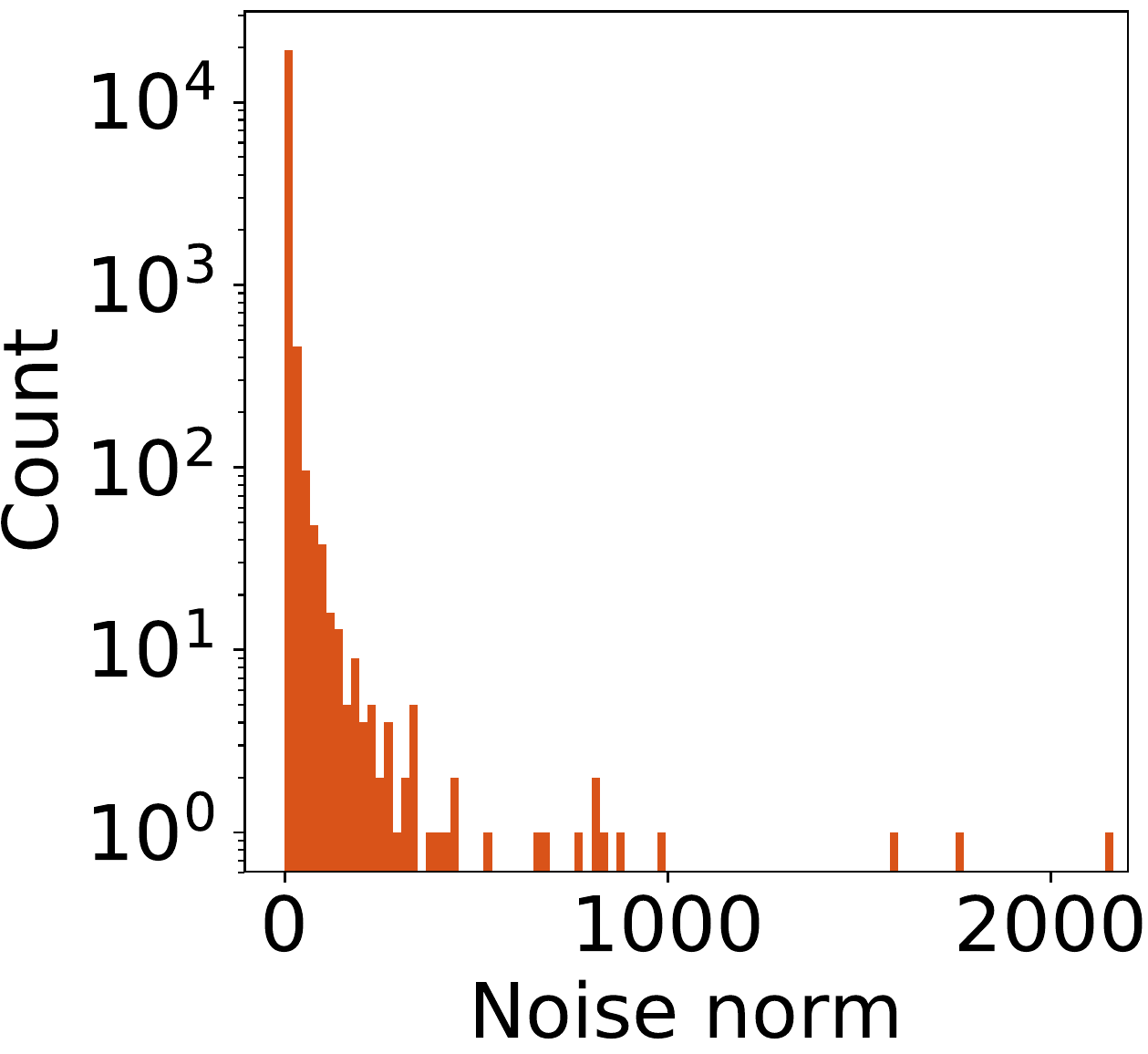} \label{fig:hist_sas}}
    \vspace{-5pt}
    \caption{(a)The histogram of the norm of the gradient noises computed with AlexNet on Cifar10. (b) and (c) the histograms of the norms of (scaled) Guassian and $\alpha$-stable random variables. }
    \vspace{-10pt}
    \label{fig:noise_norms}
\end{figure}

We first illustrate a typical mismatch between the Gaussianity assumption and the empirical behavior of the stochastic gradient noise.
In Figure~\ref{fig:noise_norms}, we plot the histogram of the norms of the stochastic gradient noise that is computed using a convolutional neural network in a real classification problem and compare it to the histogram of the norms of Gaussian random variables. It can be clearly observed that the shape of the real histogram is very different than the Gaussian and shows a \emph{heavy-tailed} behavior.

In addition to the empirical observations, the Gaussianity assumption also yields some theoretical issues. The first issue with this assumption is that the current SDE analyses of SGD are based on the \emph{invariant measure} of the SDE, which implicitly assumes that sufficiently many iterations have been taken to converge to that measure. Recent results on ULA \cite{raginsky17a,xu2018global} have shown that, the required number of iterations to achieve the invariant measure often grows exponentially with the dimension $p$. This result contradicts with the current practice: considering the large size of the neural networks and limited computational budget, only a limited number of iterations -- which is much smaller than $\exp(\mathcal{O}(p))$ -- can be taken. This conflict becomes clearer in the light of the recent works that studied the \emph{local} behavior of ULA \cite{tzen2018local,zhang17b}. These studies showed that ULA will get close to the nearest local optimum in polynomial time; however, the required amount of time for escaping from that local optimum increases exponentially with the dimension. Therefore, the phenomenon that SGD prefers wide minima within a considerably small number of iterations cannot be explained using the asymptotic distribution of the SDE given in \eqref{eqn:sgd_langevin}.

The second issue is related to the local behavior of the process and becomes clear when we consider the \emph{metastability} analysis of Brownian motion-driven SDEs. These studies \cite{freidlin1998random,bovier2004metastability,Imkeller2010} consider the case where $\wb_0$ is initialized in a quadratic basin and then analyze the minimum time $t$ such that $\wb_t$ is outside that basin. They show that this so-called \emph{first exit time} depends \emph{exponentially} on the height of the basin; however, this dependency is only \emph{polynomial} with the width of the basin. These theoretical results directly contradict with the the wide minima phenomenon: even if the height of a basin is slightly larger, the exit-time from this basin will be dominated by its height, which implies that the process would stay longer in (or in other words, `prefer') deeper minima as opposed to wider minima. The reason why the exit-time is dominated by the height is due to the \emph{continuity} of the Brownian motion, which is in fact a direct consequence of the Gaussian noise assumption.

A final remark on the issues of this approach is the observation that landscape is flat at the bottom regardless of the batch size used in SGD \cite{sagun2017empirical}. In particular, the spectrum of the Hessian at a near critical point with close to zero loss value has many near zero eigenvalues. Therefore, local curvature measures that are used as a proxy for measuring the width of a basin correlates with the magnitudes of large eigenvalues of the Hessian which are few. Besides, during the dynamics of SGD it has been observed that the algorithm does not cross barriers except perhaps at the very initial phase \cite{xing2018walk,Baity18}. Such dependence of width on an essentially-flat landscape combined with the lack of explicit barrier crossing during the SGD descent forces us to rethink the analysis of basin hopping under a noisy dynamics.


%

\textbf{Proposed framework: }
In this study, we aim at addressing these contradictions and come up with an arguably better-suited hypothesis for the stochastic gradient noise that has more pertinent theoretical implications for the phenomena associated with SGD. In particular, we go back to \eqref{eqn:stoch_grad} and \eqref{eqn:noise_gauss} and reconsider the application of CLT. This \emph{classical} CLT assumes that $U_k$ is a sum of many independent and identically distributed (i.i.d.)\ random variables, whose variance is \emph{finite}, and then it states that the law of $U_k$ converges to a Gaussian distribution, which then paves the way for \eqref{eqn:sgd_gauss}. 
Even though the finite-variance assumption seems natural and intuitive at the first sight, it turns out that in many domains, such as turbulent motions \cite{weeks1995observation}, oceanic fluid flows \cite{woyczynski2001levy}, finance \cite{mandelbrot2013fractals}, biological evolution \cite{jourdain2012levy}, audio signals \cite{liutkus2015generalized}, the assumption might fail to hold (see \cite{duan} for more examples). In such cases, the classical CLT along with the Gaussian approximation will no longer hold. While this might seem daunting, fortunately, one can prove an \emph{extended} CLT and show that the law of the sum of these i.i.d.\ variables with infinite variance still converges to a family of \emph{heavy-tailed} distributions that is called the $\alpha$-stable distribution \cite{paul1937theorie}. As we will detail in Section~\ref{sec:levy_sde}, these distributions are parametrized by their \emph{tail-index} $\alpha \in (0,2]$ and they coincide with the Gaussian distribution when $\alpha =2$. 

In this study, we relax the finite-variance assumption on the stochastic gradient noise and by invoking the extended CLT, we assume that $U_k$ follows an $\alpha$-stable distribution, as hinted in Figure~\ref{fig:hist_sas}. By following a similar rationale to \eqref{eqn:sgd_gauss} and \eqref{eqn:sgd_langevin}, we reformulate SGD with this new assumption and consider its continuous-time limit for small step-sizes. Since the noise might not be Gaussian anymore (i.e.\ when $\alpha \neq 2$), the use of the Brownian motion would not be appropriate in this case and we need to replace it with the $\alpha$-stable L\'{e}vy motion, whose increments have an $\alpha$-stable distribution \cite{yanovsky2000levy}. Due to the heavy-tailed nature of $\alpha$-stable distribution, the L\'{e}vy motion might incur large discontinuous jumps and therefore exhibits a fundamentally different behavior than the Brownian motion, whose paths are on the contrary almost surely continuous. As we will describe in detail in Section~\ref{sec:levy_sde}, the discontinuities also reflect in the metastability properties of L\'{e}vy-driven SDEs, which indicate that, as soon as $\alpha <2$, the first exit time from a basin does \emph{not} depend on its height; on the contrary, it directly depends on its width and the tail-index $\alpha$. Informally, this implies that the process will \emph{escape} from narrow minima -- no matter how deep they are -- and stay longer in wide minima. Besides, as $\alpha$ get smaller, the probability for the dynamics to jump in a wide basin will increase.
Therefore, if the $\alpha$-stable assumption on the stochastic gradient noise holds, then the existing metastability results automatically provide strong theoretical insights for illuminating the behavior of SGD.

\textbf{Contributions: }
The main contributions of this paper are twofold: (i) we perform an extensive empirical analysis of the tail-index of the stochastic gradient noise in deep neural networks and (ii) based on these empirical results, we bring an alternative perspective to the existing approaches for analyzing SGD and shed more light on the folklore that SGD prefers wide minima by establishing a bridge between SGD and the related theoretical results from statistical physics and stochastic analysis.

We conduct experiments on the most common deep learning architectures. In particular, we investigate the tail behavior under fully-connected and convolutional models using negative log likelihood and linear hinge loss functions on MNIST, CIFAR10, and CIFAR100 datasets. 
For each configuration, we scale the size of the network and batch size used in SGD and monitor the effect of each of these settings on the tail index $\alpha$. 

Our experiments reveal several remarkable results:
\begin{itemize}[noitemsep,topsep=0pt,leftmargin=*,align=left]
\item In all our configurations, the stochastic gradient noise turns out to be highly non-Gaussian and possesses a heavy-tailed behavior.
\item Increasing the size of the minibatch has a very little impact on the tail-index, and as opposed to the common belief that larger minibatches result in Gaussian gradient noise, the noise is still far from being Gaussian.
\item There is a strong interaction between the network architecture, network size, dataset, and the tail-index, which ultimately determine the dynamics of SGD on the training surface. This observation supports the view that, the geometry of the problem and the dynamics induced by the algorithm cannot be separated from each other. 
\item In almost all configurations, we observe two distinct phases of SGD throughout iterations. During the first phase, the tail-index rapidly decreases and SGD possesses a clear jump when the tail-index is at its lowest value and causes a sudden jump in the accuracy. This behavior strengthens the view that SGD crosses barriers at the very initial phase.
\end{itemize}

Our methodology also opens up several interesting future directions and open questions, as we discuss in Section~\ref{sec:conc}.

%% file: levy_sde.tex

\section{Stable distributions and SGD as a L\'{e}vy-Driven SDE}
\label{sec:levy_sde}

{The CLT states that the sum of i.i.d. random variables with a finite second moment converges to a normal distribution if the number of summands grow. However, if the variables have heavy-tail, the second moment may not exist. For instance, if their density $p(x)$ has a power-law tail decreasing as $1/|x|^{\alpha+1}$ where $0 < \alpha < 2$; only $\alpha$-th moment exist with $\alpha<2$. In this case, generalized central limit theorem (GCLT) says that the sum of such variables will converge to a distribution called the  \emph{$\alpha$-stable} distribution instead as the number of summands grows (see e.g. \cite{fischer2010history}. In this work, we focus on the centered \emph{symmetric $\alpha$-stable} ($\sas$) distribution, which is a special case of $\alpha$-stable distributions that are symmetric around the origin.
}

\begin{figure}[t]
    \centering
    \includegraphics[width=0.48\columnwidth]{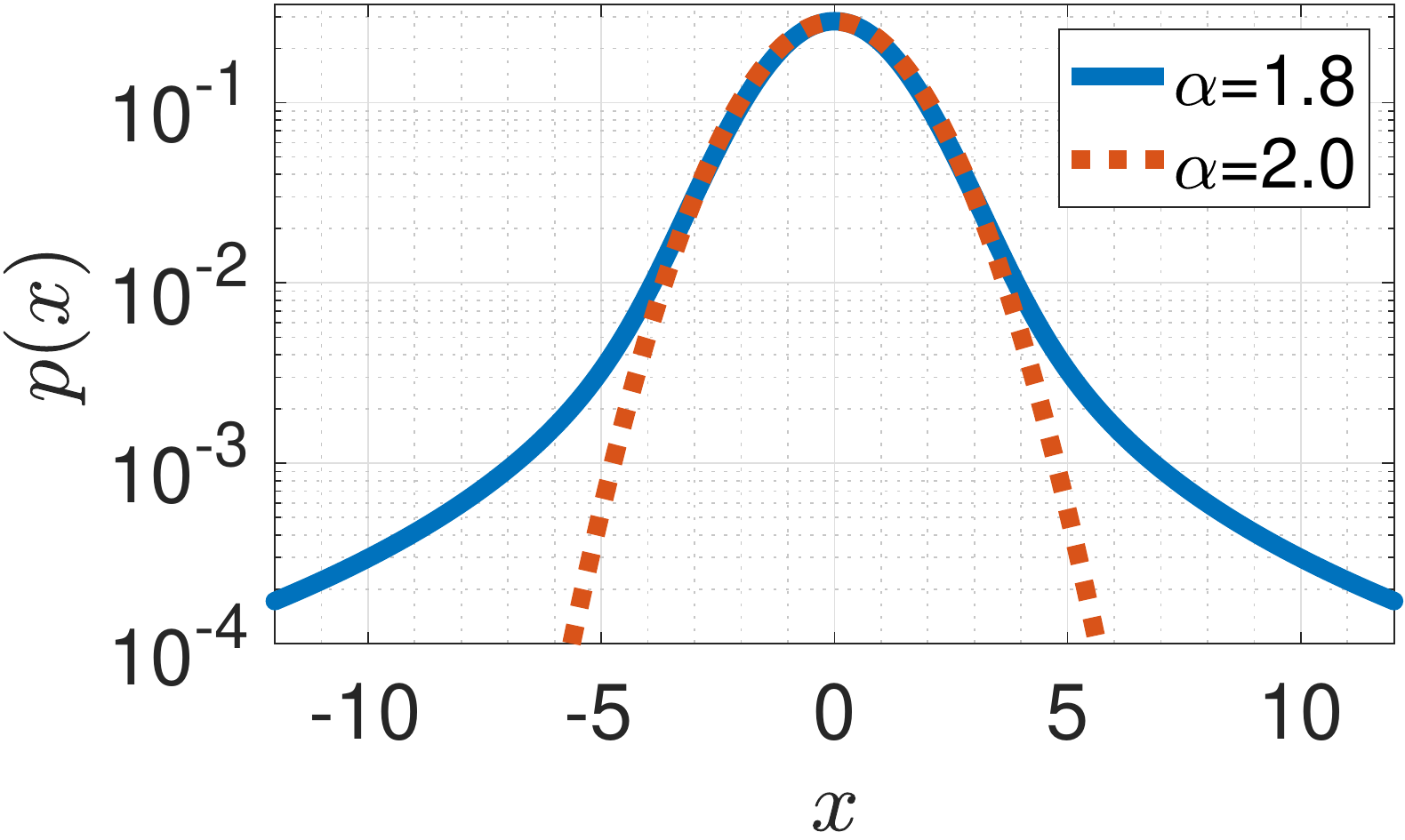}
    \includegraphics[width=0.48\columnwidth]{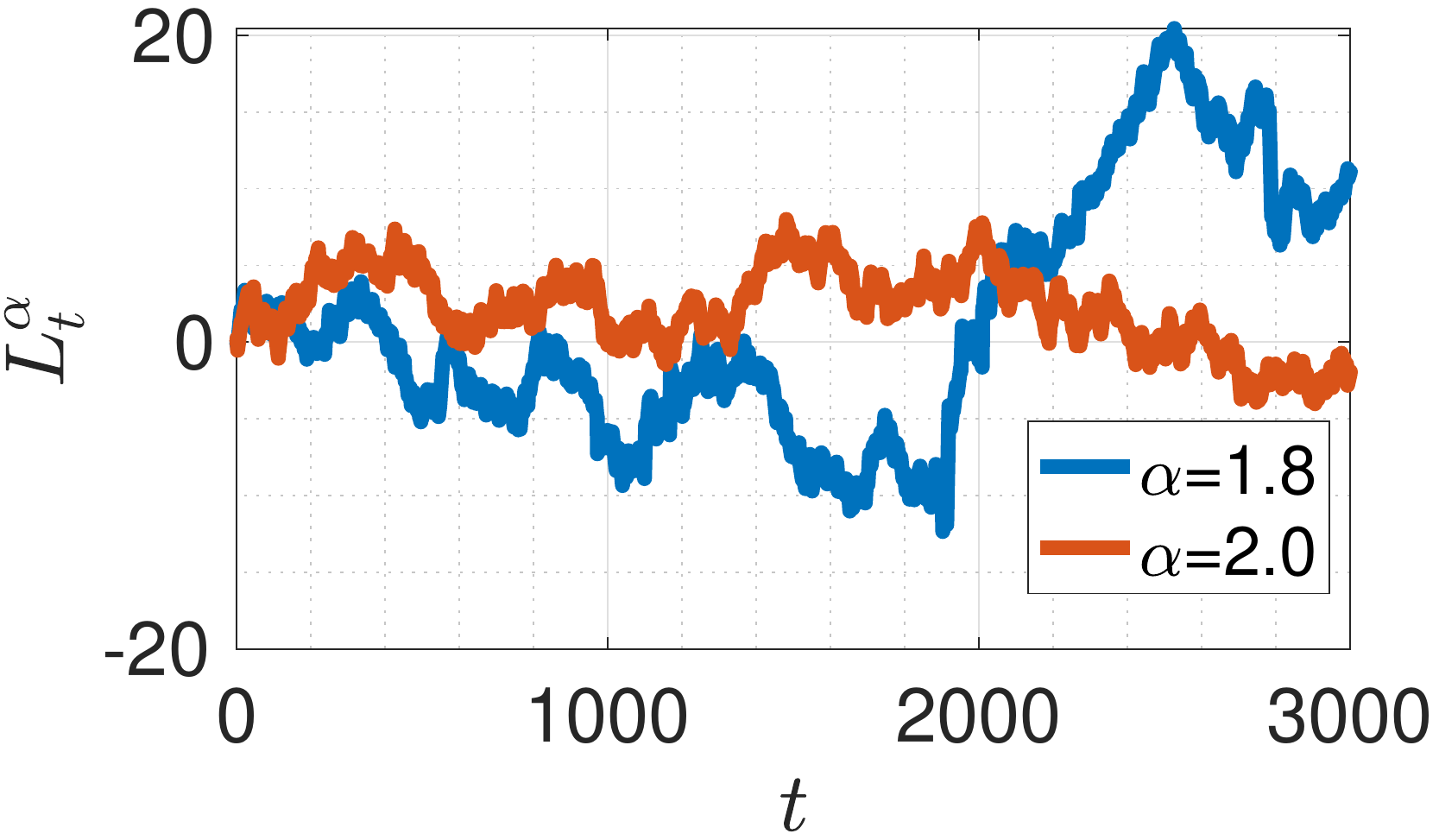}
    \caption{Left: $\sas$ densities, right:  $\mathrm{L}_t^\alpha$ for $p=1$. For $\alpha<2$, $\sas$ becomes heavier-tailed and $\mathrm{L}_t^\alpha$ incurs jumps. }
    \vspace{-10pt}
    \label{fig:sas_lm}
\end{figure}

We can view the $\sas$ distribution as a heavy-tailed generalization of a centered Gaussian distribution. The $\sas$ distributions are defined through their characteristic function via $X\sim \sas(\sigma) \iff \E[\exp(i \omega X)] = \exp(-|\sigma \omega|^\alpha)$. Even though their probability density function does not admit a closed-form formula in general except in special cases, their density decays with a power law tail like $1/|x|^{\alpha+1}$ where $\alpha \in (0,2]$ is called the \emph{tail-index} which determines the behavior of the distribution: as $\alpha$ gets smaller; the distribution has a heavier tail. In fact, the parameter $\alpha$ also determines the moments: $\mathds{E}[|X|^r] < \infty$ if and only if $r<\alpha$; implying $X$ has infinite variance when $\alpha\neq 2$. The parameter $\sigma \in \mathds{R}_+$ is known as the \emph{scale} parameter and controls the spread of $X$ around $0$. We recover the Gaussian distribution ${\cal N}(0,2\sigma^2)$ as a special case of $\sas$ when $\alpha=2$.

In this study, we make the following assumption on the stochastic gradient noise:
\begin{align}
[U_k(\wb)]_i \sim \sas(\sigma(\wb)), \quad \forall i =1,\dots,n \label{eqn:noise_levy} 
\end{align}
where $[v]_i$ denotes the $i$'th component of a vector $v$. Informally, we assume that each coordinate of $U_k$ is $\sas$ distributed with the same $\alpha$ and the scale parameter $\sigma$ depends on the state $\wb$. Here, this dependency is not crucial since we are mainly interested in the tail-index $\alpha$, which can be estimated \emph{independently} from the scale parameter. Therefore, we will simply denote $\sigma(\wb)$ as $\sigma$ for clarity.

By using the assumption \eqref{eqn:noise_levy}, we can rewrite the SGD recursion as follows:
\begin{align}
\wb_{k+1} = \wb_{k} - \eta \nabla f(\wb_k) + \eta^{1/\alpha} \Bigl(\eta^{\frac{\alpha-1}{\alpha} } \sigma\Bigr) S_k, \label{eqn:sgd_alpha}
\end{align}
where $S_k \in \mathbb{R}^p$ is a random vector such that $[S_k]_i \sim \sas(1)$. If the step-size $\eta$ is small enough, then we can consider the continuous-time limit of this discrete-time process, which is expressed in the following SDE driven by an $\alpha$-stable L\'{e}vy process:
\begin{align}
\rmd \wb_t = - \nabla f(\wb_t) \rmd t + \eta^{(\alpha-1)/\alpha} \sigma \> \rmd \Lm_t, \label{eqn:sgd_levy}
\end{align}
where $\Lm_t$ denotes the $p$-dimensional $\alpha$-stable L\'{e}vy motion with \emph{independent components}. In other words, each component of $\Lm_t$ is an independent $\alpha$-stable L\'{e}vy motion in $\mathbb{R}$. For the scalar case it is defined as follows for $\alpha \in (0,2]$ \cite{duan}:

\begin{enumerate}[label=(\roman*),itemsep=0pt,topsep=0pt,leftmargin=*,align=left]
\item $\Lm_0 = 0$ almost surely.
\item For $t_0<t_1 < \cdots < t_N$, the increments $ (\Lm_{t_{i}} - \Lm_{t_{i-1}} )$ are independent ($i = 1,\dots, N$). 
\item The difference $(\Lm_t - \Lm_s)$ and $\Lm_{t-s}$ have the same distribution: $\sas((t-s)^{1/\alpha})$ for $s<t$. 
\item $\Lm_t$ is continuous in probability (i.e.\ it has \emph{stochastically continuous} sample paths): for all $\delta >0$ and $s\geq 0$, $p(|\Lm_t - \Lm_s| > \delta) \rightarrow 0$ as $t \rightarrow s$.
\end{enumerate}
When $\alpha = 2$, $\Lm_t$ coincides with a scaled version of Brownian motion, $\sqrt{2} \Bm_t$. $\sas$ and $\Lm_t$ are illustrated in Figure~\ref{fig:sas_lm}.

The SDE in \eqref{eqn:sgd_levy} exhibits a fundamentally different behavior than the one in \eqref{eqn:sgd_langevin} does. This is mostly due to the stochastic continuity property of $\Lm_t$, which enables $\Lm_t$ to have a countable number of discontinuities, which are sometimes called `jumps'.
 In the rest of this section, we will recall important theoretical results about this SDE and discuss their implications on SGD.

For clarity of the presentation and notational simplicity we focus on the scalar case and consider the SDE \eqref{eqn:sgd_levy} in $\rset$ (i.e.\ $p=1$). Multidimensional generalizations of the metastability results presented in this paper can be found in \cite{imkeller2010first}. We rewrite \eqref{eqn:sgd_levy} as follows:
\beq \label{eq-levy-sde} \rmd  w_t^\varepsilon = -\nabla f(w_t^\varepsilon) \rmd t + \varepsilon \rmd \Lm_t  
\eeq
for $t\geq 0$, started from the initial point $w_0\in\mathbb{R}$, where $\Lm_t$ is the $\alpha$-stable L\'evy process, $\varepsilon \geq 0$ is a parameter and $f$ is a non-convex objective with $r \geq 2$ local minima. 

When $\varepsilon=0$, we recover the gradient descent dynamics in continuous time: $\rmd  w_t^0 = -\nabla f(w_t^0) \rmd t$,
%
where the local minima are the stable points of this differential equation. However, as soon as $\varepsilon >0$, these states become `metastable', meaning that there is a positive probability for $w_t^\varepsilon$ to transition from one basin to another. However, the time required for transitioning to another basin strongly depends on the characteristics of the injected noise. 
The two most important cases are $\alpha =2$ and $\alpha < 2$. When $\alpha =2$, (i.e.\ the Gaussianity assumption) the process $(w^\varepsilon_t)_{t \geq 0}$ is continuous, which requires it to `climb' the basin all the way up, in order to be able to transition to another basin. This fact makes the transition-time depend on the height of the basin. On the contrary, when $\alpha <2$, the process can incur discontinuities and do not need to cross the boundaries of the basin in order to transition to another one since it can directly jump. This property is called the `transition phenomenon' \cite{duan} and makes the transition-time mostly depend on the \emph{width} of the basin. In the rest of the section, we will formalize these explanations.

Under some assumptions on the objective $f$, it is known that the process \eqref{eq-levy-sde} admits a stationary density \cite{Samorodnitsky2003}. For a general $f$, an explicit formula for the equilibrium distribution is not known, however when the noise level $\varepsilon$ is small enough, finer characterizations of the structure of the equilibrium density in dimension one is known. We next summarize known results in this area, which show that L\'{e}vy-driven dynamics spends more time in `wide valleys' in the sense of \cite{entropy-sgd} when $\varepsilon$ goes to zero.

Assume that $f$ is smooth with $r$ local minima $\{m_i\}_{i=1}^r$ separated by $r-1$ local maxima $\{s_i\}_{i=1}^{r-1}$, i.e. 
\beqs -\infty := s_0 < m_1 < s_1  < \dots <s_{r-1} < m_r < s_r := \infty. \eeqs
Furthermore, assume that the local minima and maxima are not degenerate, i.e. $f''(m_i)>0$ and $f''(s_i)<0$ for every $i$. We also assume the objective gradient has a growth condition $f'(w) >|w|^{1+c}$ for some constant $c>0$ and when $|w|$ is large enough. Each local minima $m_i$ lies in the (interval) valley $S_i = (s_{i-1},s_i)$ of (width) length $L_i = |s_i-s_{i-1}|$. Consider also a $\delta$-neighborhood $B_i := \{ |x - m_i|\leq \delta \}$ around the local minimum with $\delta>0$ small enough so that the neighborhood is contained in the valley $S_i$ for every $i$. We are interested in the first exit time from $B_i$ starting from a point $w_0\in B_i$ and the transition time $T_{w_0}^i(\varepsilon):= \inf \{ t\geq 0 : w_t^\varepsilon \not\in \cup_{j\neq i} B_j \}$ to a neighborhood of another local minimum, we will remove the dependency to $w_0$ of the transition time in our discussions as it is clear from the context. The following result shows that the transition times are asymptotically exponentially distributed in the limit of small noise and scales like $\frac{1}{\varepsilon^\alpha}$ with $\varepsilon$.
 \begin{thm}[\cite{pavlyukevich2007cooling}]\label{thm-levy-exit} For an initial point $w_0\in B_i$, in the limit $\varepsilon\to0$, the following statements hold regarding the transition time:
   \beqs 
         \mathbb{P}_{w_0}(T^i(\varepsilon) \in B_j) &\to& q_{ij} q_i^{-1} \quad \mbox{if} \quad i\neq j,  \\
         \mathbb{P}_{w_0}(\varepsilon^\alpha T^i(\varepsilon) \geq u ) &\leq& e^{-q_i u} \quad \mbox{for any} \quad u\geq 0.
  \eeqs
 where
  \beq    
         q_{ij} &=& \frac{1}{\alpha} \left| \frac{1}{|s_{j-1} - m_i |^\alpha} - \frac{1}{|s_{j} - m_i |^\alpha} \right|, \\
         q_{i} &=& \sum_{j\neq i}q_{ij}. 
     \eeq
 \end{thm}
 If the SDE \eqref{eq-levy-sde} would be driven by the Brownian motion instead, then an analogous theorem to Theorem \ref{thm-levy-exit} holds saying that the transition times are still exponentially distributed but the scaling $\varepsilon^\alpha$ needs to be replaced by $e^{2H/\varepsilon^2}$ where $H$ is the maximal depth of the basins to be traversed between the two local minima \cite{Day83,bovier2005metastability}. This means that in the small noise limit, Brownian-motion driven gradient descent dynamics need exponential time to transit to another minimum whereas Levy-driven gradient descent dynamics need only polynomial time. We also note from Theorem \ref{thm-levy-exit} that the mean transition time between valleys for L\'evy SDE does not depend on the depth $H$ of the valleys they reside in which is an advantage over Brownian motion driven SDE in the existence of deep valleys. Informally, this difference is due to the fact that Brownian motion driven SDE has to typically climb up a valley to exit it, whereas L\'{e}vy-driven SDE could jump out. 

\begin{figure}[t]
    \centering
    \subfigure[]{\label{fig:double_well} \includegraphics[width=0.49\columnwidth]{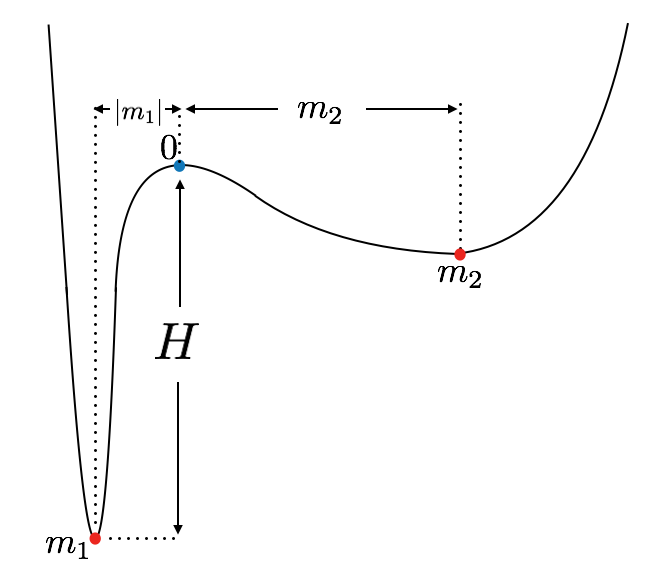}}
    \subfigure[]{\label{fig:exp_synth}   \includegraphics[width=0.45\columnwidth]{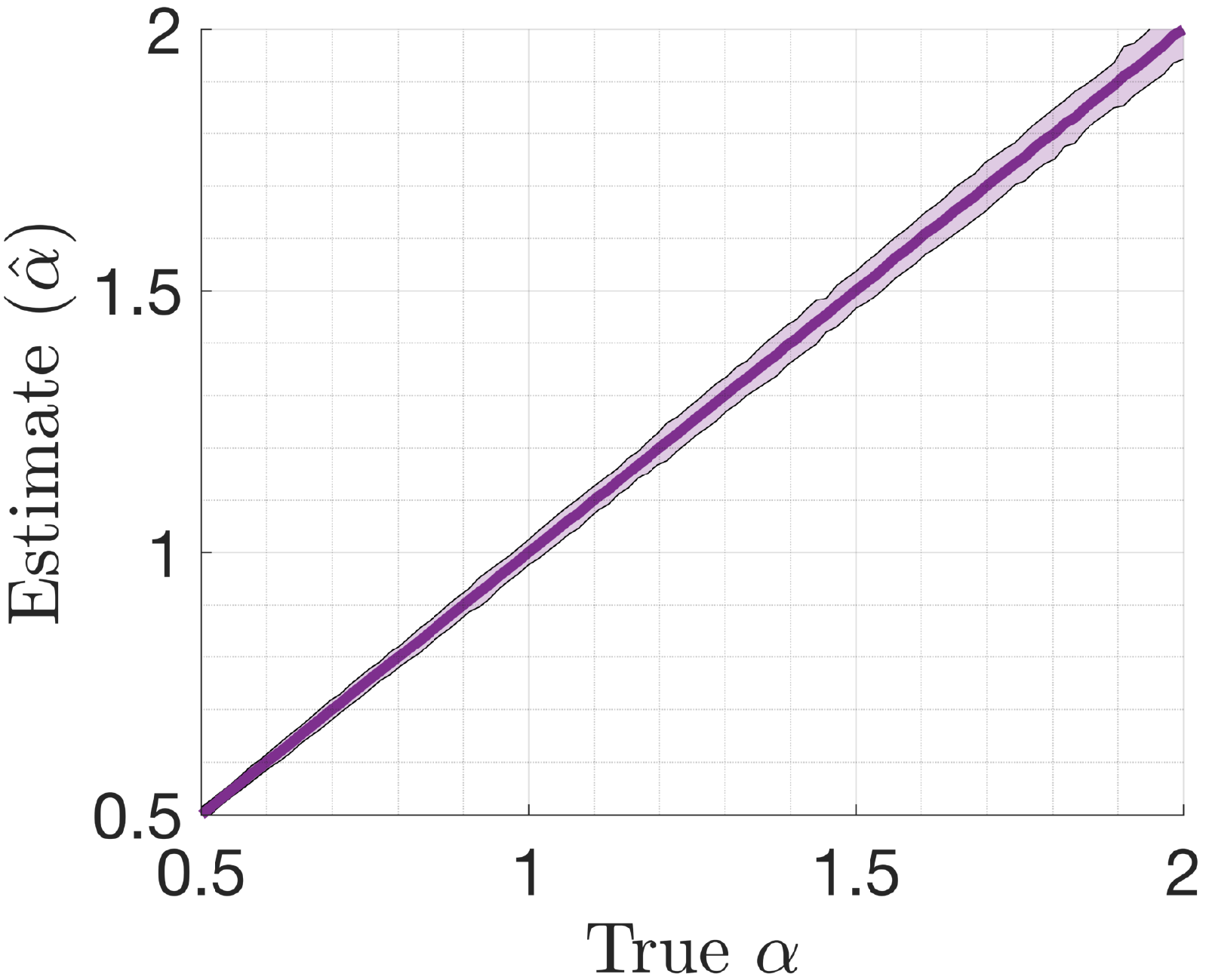}}
    \vspace{-5pt}
    \caption{ (a) An objective with two local minima $m_1, m_2$ seperated by a local maxima at $s_1 = 0$. (b) Illustration of the tail-index estimator $\hat{\alpha}$. 
    }
    \vspace{-5pt}
\end{figure}

The following theorem says that as $\varepsilon \to 0$, up to a normalization in time, the process $w_t^\varepsilon$ behaves like a finite state-space Markov process that has support over the set of local minima $\{m_i\}_{i=1}^r$ admitting a stationary density $\pi = (\pi_i)_{i=1}^r$ with an infinitesimal generator $Q$. The process jumps between the valleys $S_i$, spending time proportional to probability $p_i$ amount of time in each valley in the equilibrium where the probabilities $\pi = (\pi_i)_{i=1}^r$ are given by the solution to the linear system $Q\pi = 0$. 
\begin{thm}[{\cite{pavlyukevich2007cooling}}]\label{thm-levy-exit} Let $w_0 \in S_i$, for some $ 1\leq i \leq r$. For $t\geq 0$, 
   $w_{t\varepsilon^{-\alpha}}^\varepsilon \to Y_{m_i}(t)$, as $\varepsilon\to 0$,
in the sense of finite-dimensional distributions, where $Y = (Y_{y}(t))_{t\geq 0}$ is a continuous-time Markov chain on a state space $\{m_1,m_2,\dots,m_r\}$ with the infinitesimal generator $Q = (q_{ij})_{i,j=1}^r$ with 
     \beq    
         q_{ij} &=& \frac{1}{\alpha} \left| \frac{1}{|s_{j-1} - m_i |^\alpha} - \frac{1}{|s_{j} - m_i |^\alpha} \right|, \\
         q_{ii} &=&-\sum \nolimits_{j\neq i} q_{ij}. 
     \eeq
This process admits a density $\pi$ satisfying $Q^T\pi = 0$.    
\end{thm}
A consequence of this theorem is that equilibrium probabilities $p_i$ are typically larger for ``wide valleys". To see this consider the special case illustrated in Figure \ref{fig:double_well} with $r=2$ local minima $m_1 < s_1 = 0 < m_2$ separated by a local maximum at $s_1 = 0$. For this example, $m_2 > |m_1|$, and the second local minimum lies in a wider valley. A simple computation reveals
 $$ \pi_1 = \frac{|m_1|^\alpha}{|m_1|^\alpha + m_2^\alpha}, \quad \pi_2 = \frac{|m_2|^\alpha}{|m_1|^\alpha + |m_2|^\alpha} $$

We see that $\pi_2 > \pi_1$, that is in the equilibrium the process spends more time on the wider walley. In particular,  the ratio $\frac{\pi_2}{\pi_1} = \left(\frac{m_2}{|m_1|}\right)^\alpha$
grows with an exponent $\alpha$ when the ratio $\frac{m_2}{|m_1|}$ of the width of the valleys grows. 
Consequently, if the gradient noise is indeed $\alpha$-stable distributed, these results directly provide theoretical evidence for the wide-minima behavior of SGD.

%% file: experiments.tex

\section{Experimental Setup and Methodology}
\label{sec:exps}

\textbf{Experimental setup: }
We investigate the tail behavior of the stochastic gradient noise in a variety of scenarios. We first consider a fully-connected network (FCN) on the MNIST and CIFAR10 datasets. For this model, we vary the depth (i.e.\ the number of layers) in the set $\{2,3,\dots,10\}$, the width (i.e.\ the number of neurons per layer) in the set $\{2,4,8,\dots,1024\}$, and the minibatch size ranging from $1$ to full batch. 
%
%
We then consider a convolutional neural network (CNN) architecture (AlexNet) on the CIFAR10 and CIFAR100 datasets. We scale the number of filters in each convolutional layer in range $\{2,4,\dots,512\}$. We randomly split the MNIST dataset into train and test parts of sizes $60$K and $10$K, and CIFAR10 and CIFAR100 datasets into train and test parts of sizes $50$K and $10$K, respectively. The order of the total number of parameters $p$ range from several thousands to tens of millions.


For both fully connected and convolutional settings, we run each configuration with the negative-log-likelihood (i.e.\ cross entropy) and with the linear hinge loss, and we repeat each experiment with three different random seeds. The training algorithm is SGD with no explicit modification such as momentum or weight decay. The training runs until 100\% training accuracy is achieved or until maximum number of iterations limit is reached (the latter limit is effective in the under-parametrized models). At every $100$th iteration, we log the full training and test accuracies, and the tail estimate of the gradients that are sampled using the corresponding mini-batch size. The codebase is implemented in python using pytorch and provided it in the supplementary material. Total runtime is $\sim$3 weeks on 8 relatively modern GPUs.

\textbf{Method for tail-index estimation:}
Estimating the tail-index of an extreme-value distribution is a long-standing topic. Some of the well-known estimators for this task are \cite{hill1975simple,pickands1975statistical,dekkers1989moment,de1998comparison}. Despite their popularity, these methods are not specifically developed for $\alpha$-stable distributions and it has been shown that they might fail for estimating the tail-index for $\alpha$-stable distributions \cite{mittnik1996tail,paulauskas2011once}. 

In this study, we use a relatively recent estimator proposed in \cite{mohammadi2015estimating} for $\alpha$-stable distributions. It is given in the following theorem. 
\begin{thm}[\cite{mohammadi2015estimating}]
Let $\{X_i\}_{i=1}^K$ be a collection of random variables with $X_i \sim \sas(\sigma)$ and $K = K_1 \times K_2$.
Define $Y_i \triangleq \sum_{j=1}^{K_1} X_{j+(i-1)K_1} \>$ for $i \in \llbracket 1, K_2 \rrbracket$. Then, the estimator
\begin{align}
\label{eqn:alpha_estim}
\widehat{\phantom{a}\frac1{\alpha}\phantom{a}} \hspace{-4pt} \triangleq \hspace{-2pt} \frac1{\log K_1} \Bigl(\frac1{K_2 } \sum_{i=1}^{K_2} \log |Y_i|  - \frac1{K} \sum_{i=1}^K \log |X_i| \Bigr). 
\end{align}
converges to $1/{\alpha}$ almost surely, as $K_2 \rightarrow \infty$.
\end{thm}
As shown in Theorem 2.3 of \cite{mohammadi2015estimating}, this estimator admits a provably faster convergence rate and smaller asymptotic variance than all the aforementioned methods.

In order to verify the accuracy of this estimator, we conduct a preliminary experiment, where we first generate $K = K_1 \times K_2$ many $\sas(1)$ distributed random variables with $K_1= 100$, $K_2 =1000$ for $100$ different values of $\alpha$. 
Then, we estimate $\alpha$ by using $\hat{\alpha} \triangleq (\widehat{\phantom{a}\frac1{\alpha}\phantom{a}})^{-1}$. We repeat this experiment $100$ times for each $\alpha$. As shown in Figure~\ref{fig:exp_synth}, the estimator is very accurate for a large range of $\alpha$. Due to its favorable theoretical properties such as independence of the scale parameter $\sigma$, combined with its empirical stability, we choose this estimator in our experiments.

In order to estimate the tail-index $\alpha$ at iteration $k$, we first partition the set of data points $\mathcal{D} \triangleq \{1,\dots,n\}$ into many disjoint sets $\Omega_k^{i} \subset \mathcal{D}$ of size $b$, such that the union of these subsets give all the data points. Formally, for all $i,j =1,\dots, n/b$, $|\Omega_k^i| = b$, $\cup_{i} \Omega_k^i = \mathcal{D}$, and $\Omega_k^i \cap \Omega_k^j=\emptyset$ for $i \neq j$. This approach is similar to sampling without replacement. We then compute the full gradient $\nabla f(\wb_k)$ and the stochastic gradients $\nabla \tilde{f}_{\Omega_k^i}(\wb_k)$ for each minibatch $\Omega_k^i$. We finally compute the stochastic gradient noises $U^i_k(\wb_k) = \nabla \tilde{f}_{\Omega_k^i}(\wb_k) - \nabla f(\wb_k)$, vectorize each $U^i_k(\wb_k)$ and concatenate them to obtain a single vector, and compute the reciprocal of the estimator \eqref{eqn:alpha_estim}. In this case, we have $K=pn/b$ and we set $K_1$ to the divisor of $K$ that is the closest to $\sqrt{K}$.

\section{Results}

In this section we present the most important and representative results. We have observed that, in all configurations, the choice of the two loss functions and the three different initializations yield no significant difference. Therefore, throughout this section, we will focus on the negative-log-likelihood loss. Unless stated otherwise, we set the minibatch size $b=500$ and the step-size $\eta = 0.1$.

\begin{figure}[t]
    \centering
    \subfigure[MNIST]{
    \includegraphics[width=0.39\columnwidth]{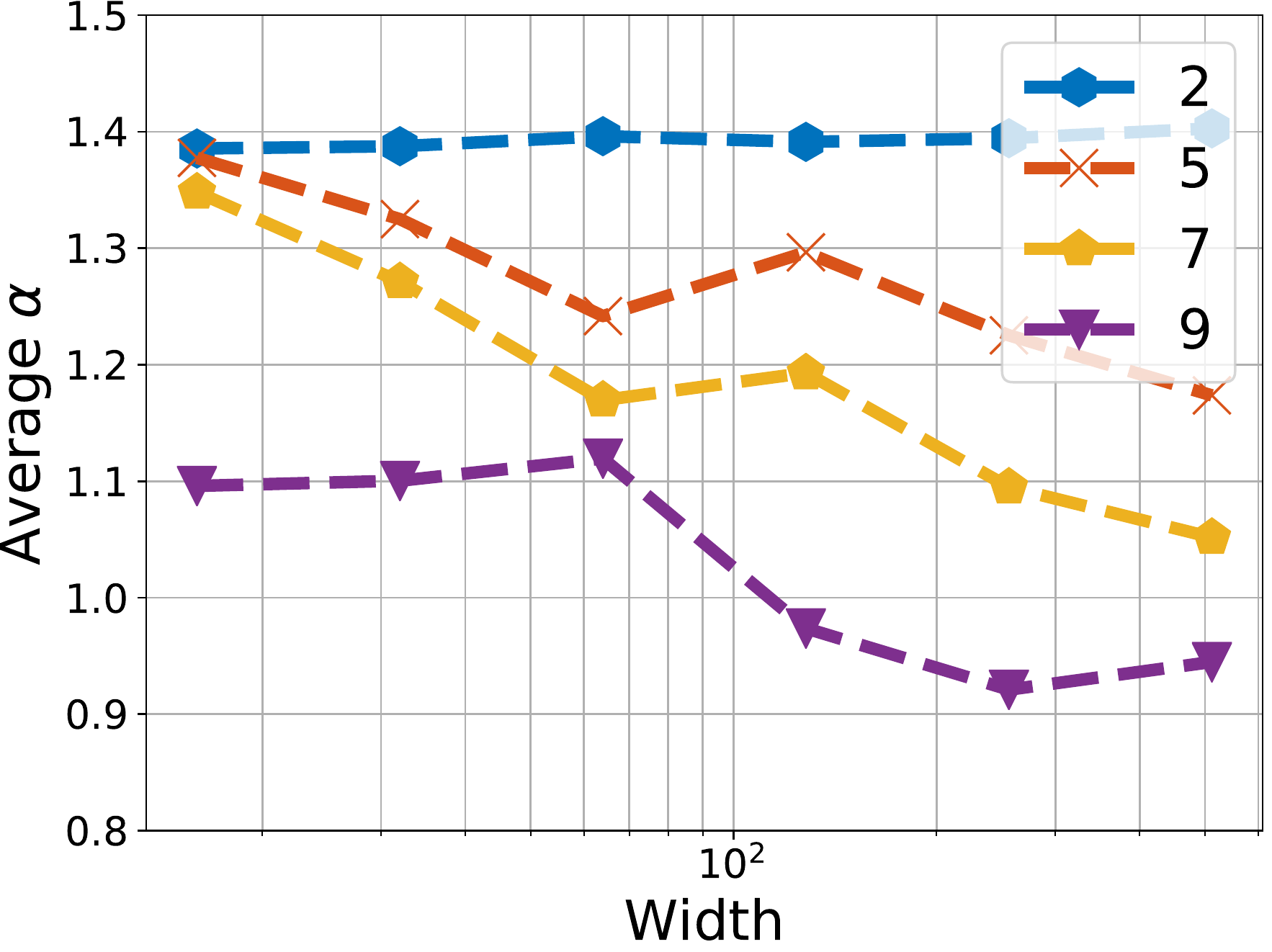}
    \includegraphics[width=0.39\columnwidth]{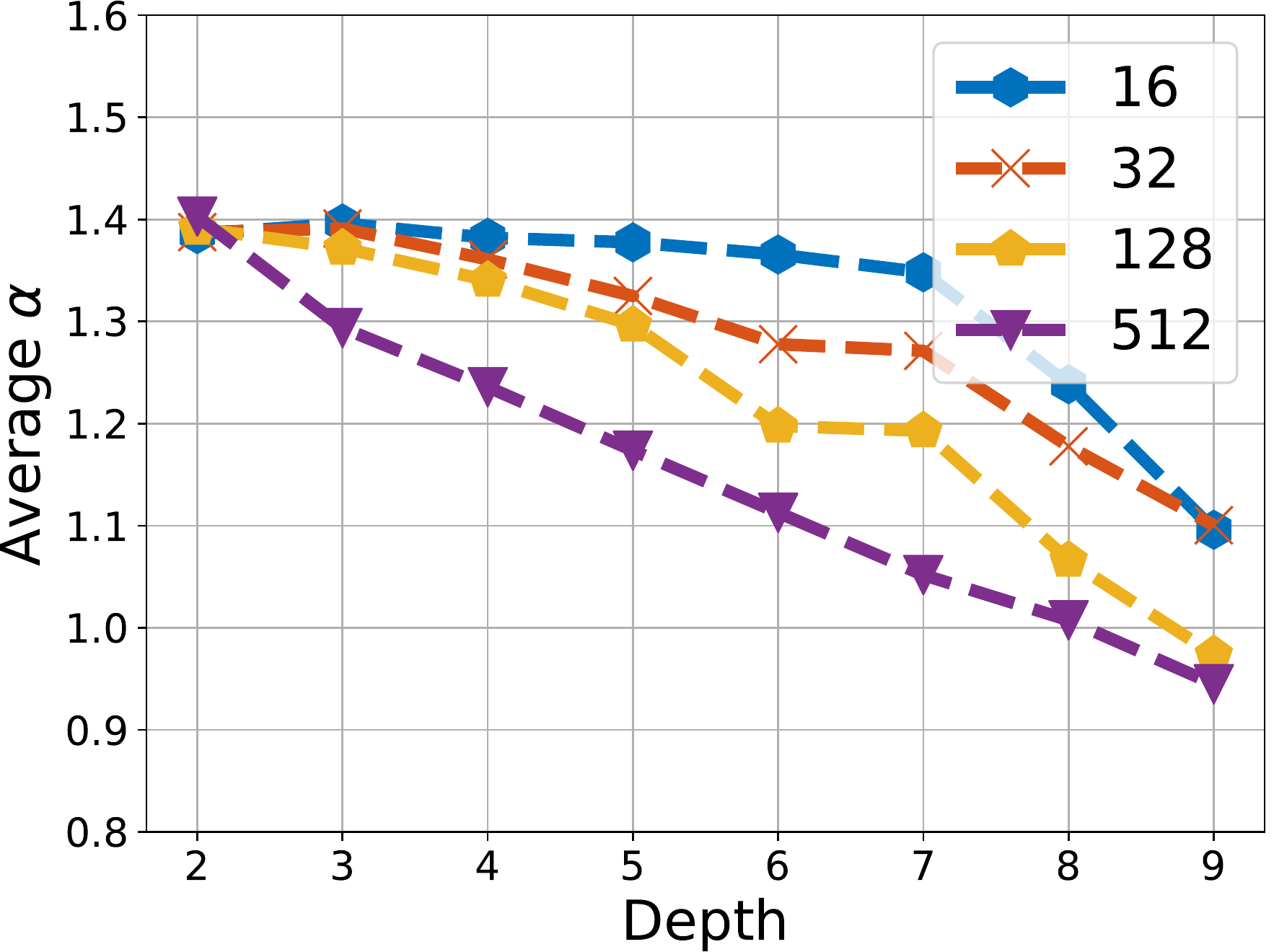}
    }
    \subfigure[CIFAR10]{\includegraphics[width=0.39\columnwidth]{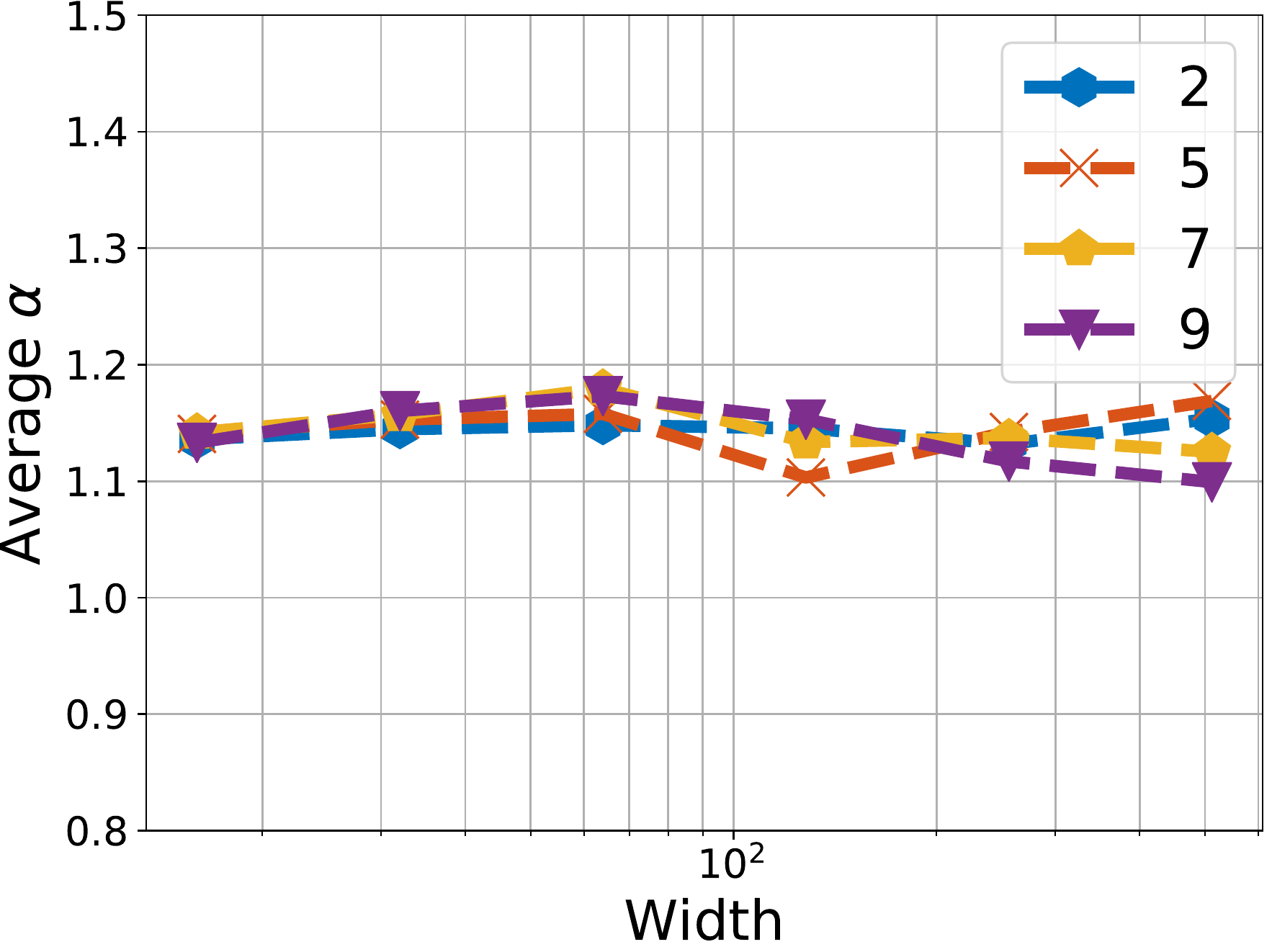}
    \includegraphics[width=0.39\columnwidth]{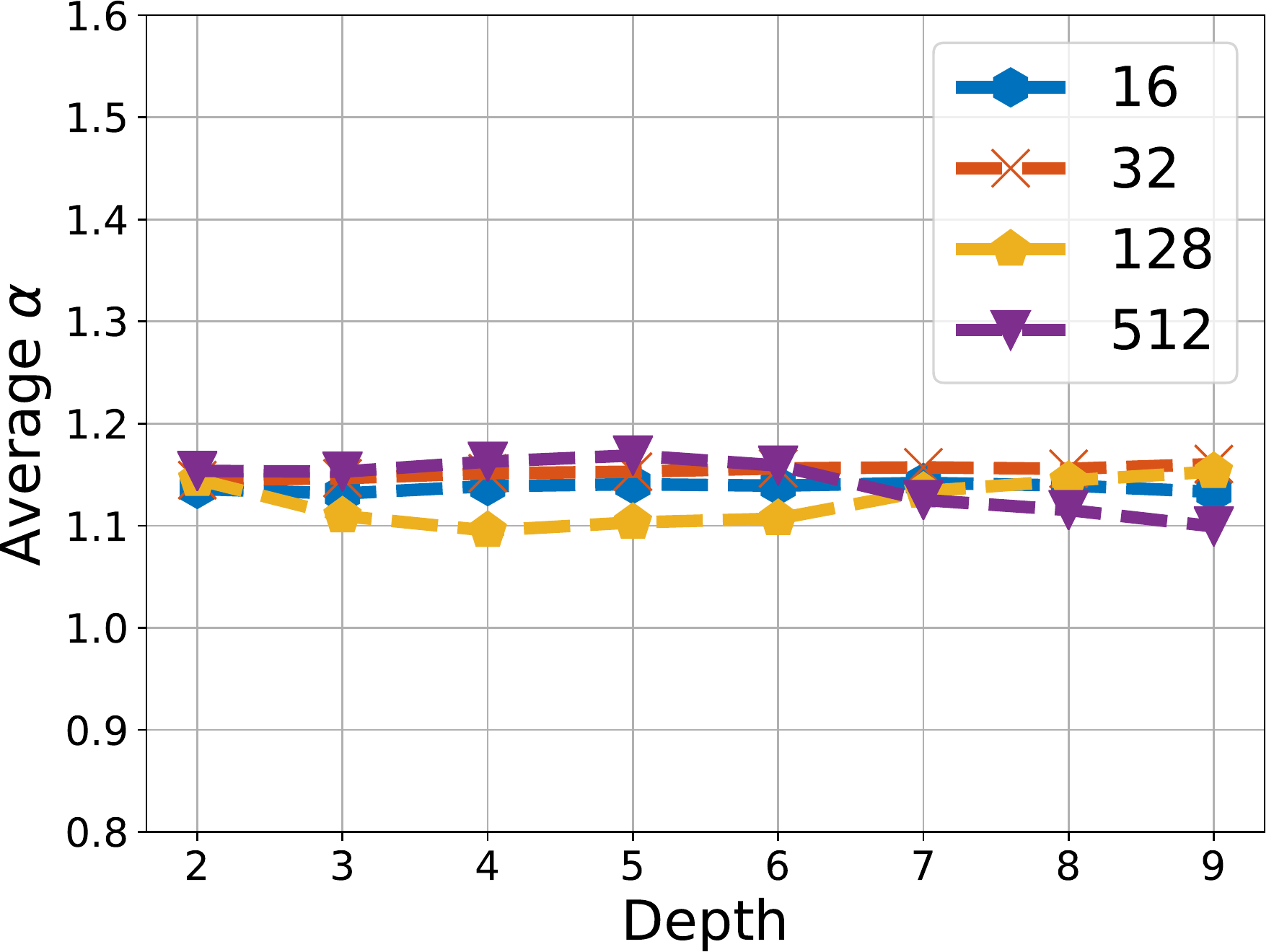}
    }
    \vspace{-15pt}
    \caption{Estimation of $\alpha$ for varying widths and depths in FCN. The curves in the left figures correspond to different depths, and the ones on the right figures correspond to widths.}
    
    \label{fig:exp_fc_widths}
\end{figure}

\textbf{Effect of varying network size: }
In our first set of experiments, we measure the tail-index for varying the widths and depths for the FCN, and varying widths (i.e.\ the number of filters) for the CNN. For very small sizes, the networks perform poorly, therefore, we only illustrate sufficiently large network sizes, which yield similar accuracies. For these experiments, we compute the average of the tail-index measurements for the last $10$K iterations (i.e.\ when $\hat{\alpha}$ becomes stationary) to focus on the late stage dynamics.

Figure~\ref{fig:exp_fc_widths} shows the results for the FCN. The first striking observation is that in all the cases, the estimated tail-index is far from $2$ with a very high confidence (the variance of the estimates were around $0.001$), meaning that the distribution of the gradient noise is highly non-Gaussian. For the MNIST dataset, we observe that $\alpha$ systematically decreases for increasing network size, where this behavior becomes more prominent with the depth. This result shows that, for MNIST, increasing the dimension of the network results in a gradient noise with heavier tails and therefore increases the probability to end up in a wider basin. 

For the CIFAR10 dataset, we still observe that $\alpha$ is far from $2$; however, in this case, increasing the network size does not have a clear effect on $\alpha$: in all cases, we observe that $\alpha$ is in the range $1.1$--$1.2$.  

\begin{figure}[t!]
    \centering
    \subfigure[CIFAR10 ]{
                    \includegraphics[width=0.39\columnwidth]{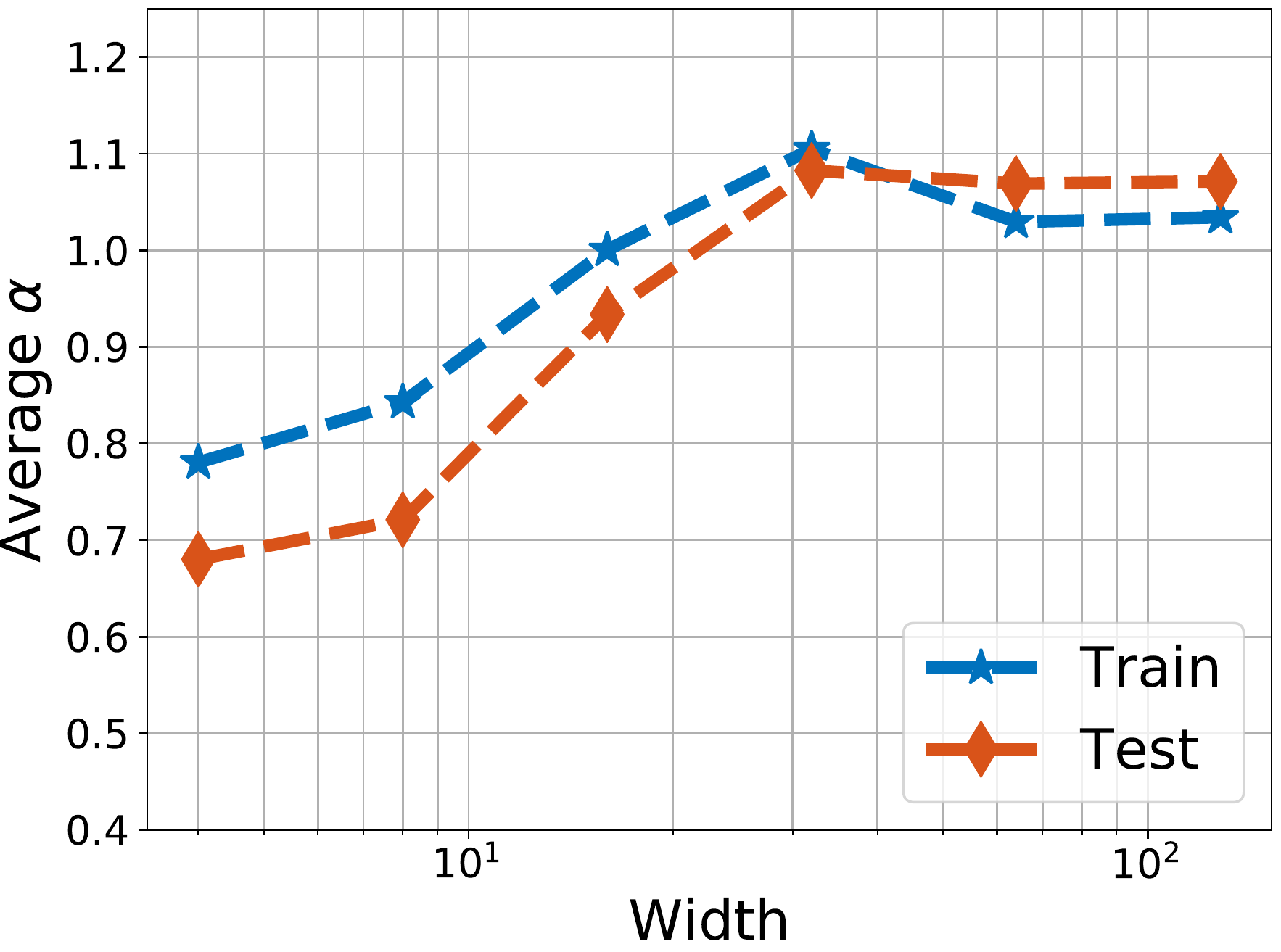}
                    \includegraphics[width=0.39\columnwidth]{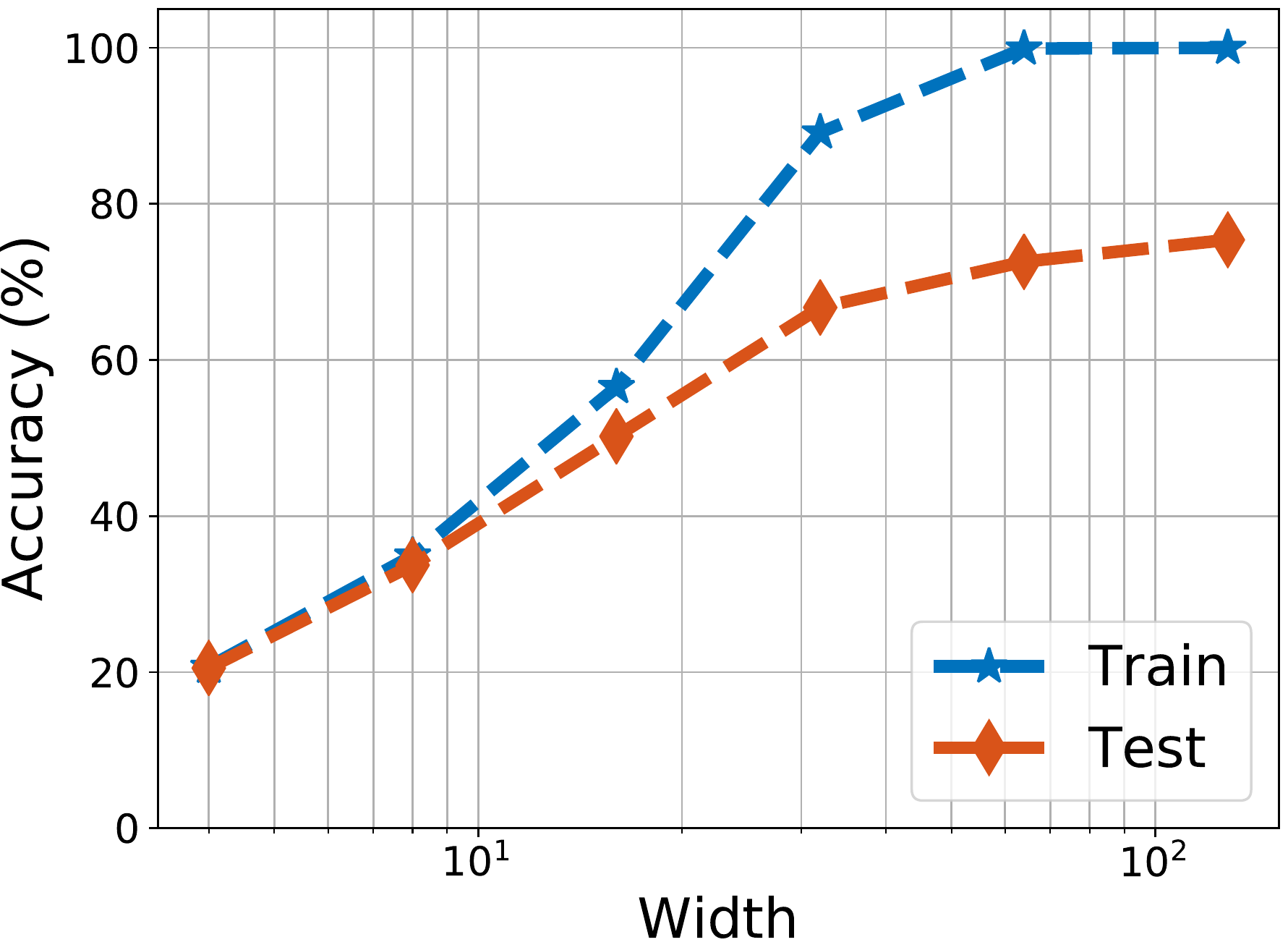}}
    \subfigure[CIFAR100 ]{
                    \includegraphics[width=0.39\columnwidth]{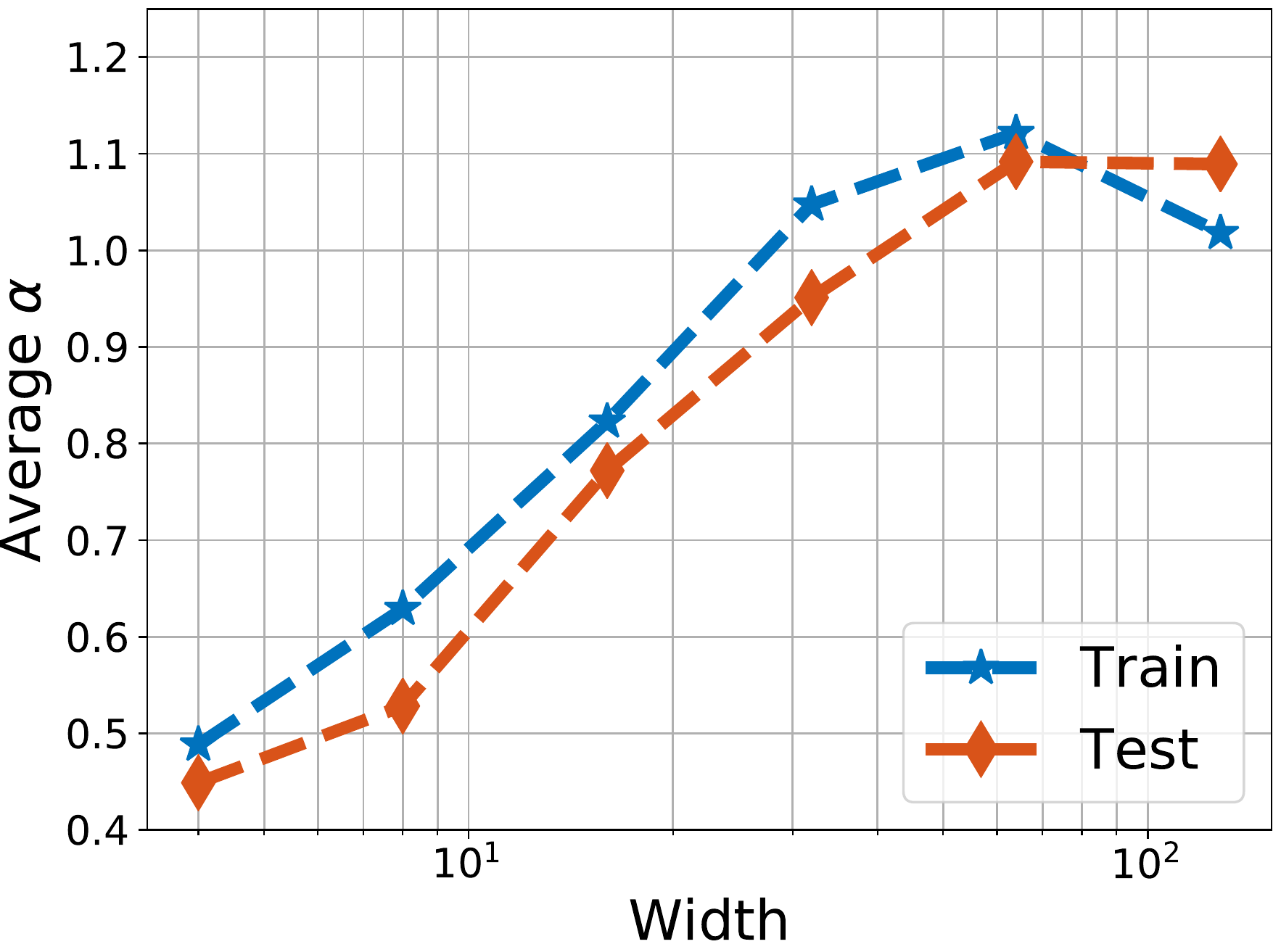}
                    \includegraphics[width=0.39\columnwidth]{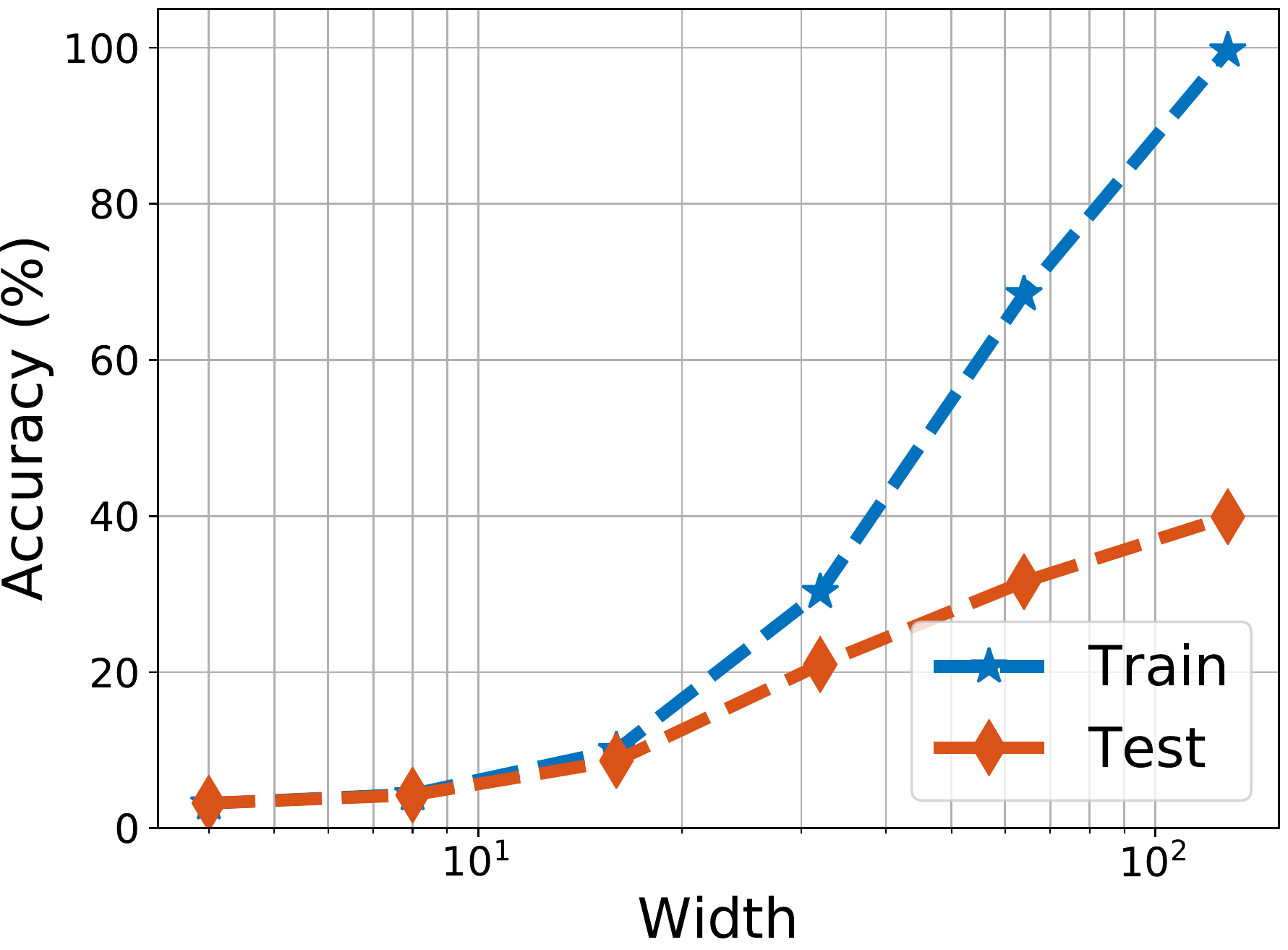}}
    \vspace{-15pt}
    \caption{The accuracy and $\hat{\alpha}$ of the CNN for varying widths.   }
    \label{fig:cnn_widths}
\end{figure}

Figure~\ref{fig:cnn_widths} shows the results for the CNN. In this figure, we also depict the train and test accuracy, as well as the tail-index that is estimated on the test set. These results show that, for both CIFAR10 and CIFAR100, the tail-index is extremely low for the under-parametrized regime (e.g.\ the case when the width is $2$, $4$, or $8$ for CIFAR10). As we increase the size of the network the value of $\alpha$ increases until the network performs reasonably well and stabilizes in the range $1.0$--$1.1$. We also observe that $\alpha$ behaves similarly for both train and test sets\footnote{We observed a similar behavior in under-parametrized FCN; however, did not plot those results to avoid clutter.}.

These results show that there is strong interplay between the network architecture, dataset, and the algorithm dynamics: (i) we see that the size of the network can strongly influence $\alpha$, (ii) for the exact same network architecture, the choice of the dataset has a significant impact on not only the landscape of the problem, but also the noise characteristics, hence on the algorithm dynamics.



\begin{figure}[t]
    \centering
    \subfigure[Depth = 2]{\includegraphics[width=0.39\columnwidth]{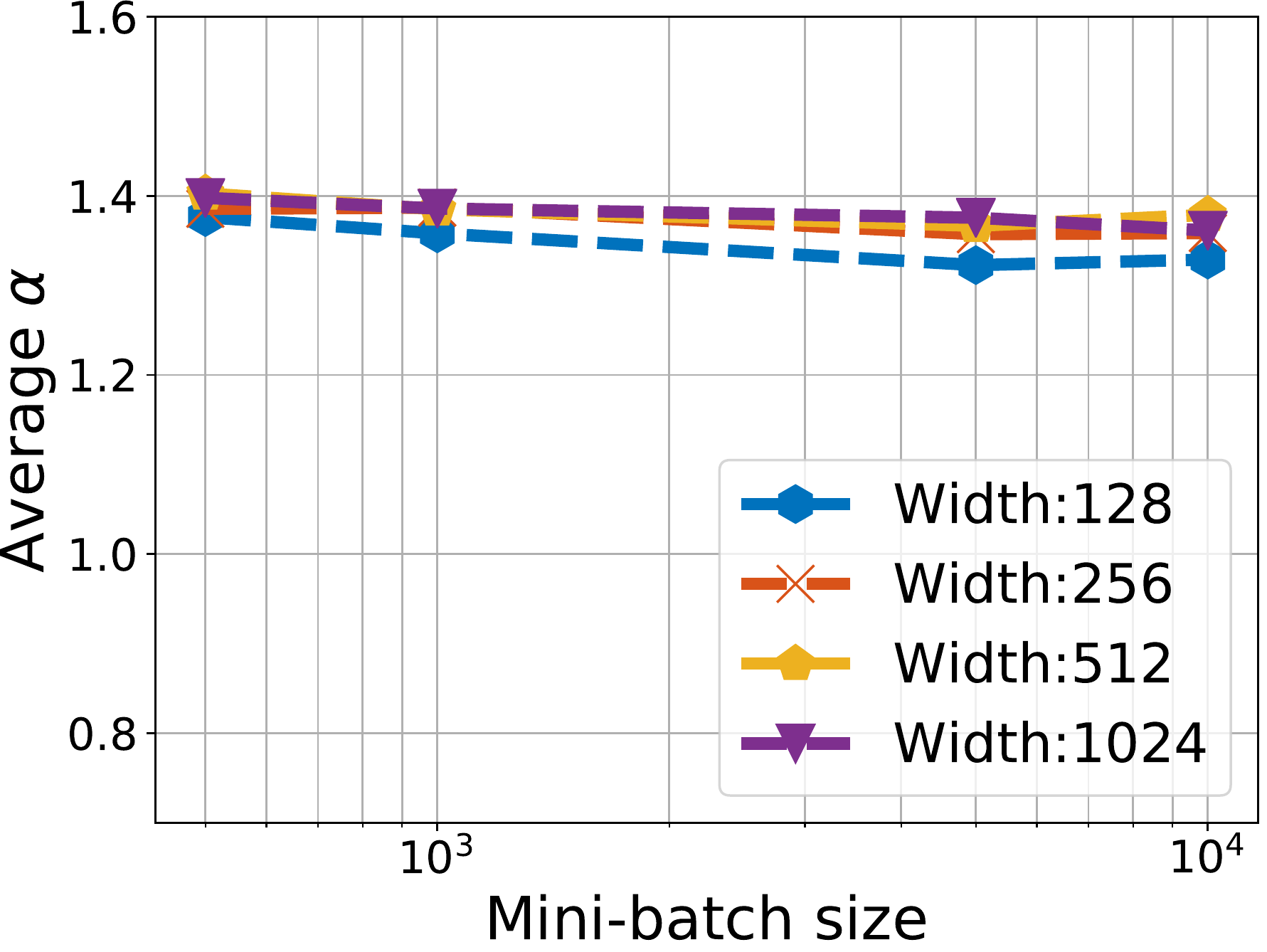}}
    \subfigure[Depth = 4]{\includegraphics[width=0.39\columnwidth]{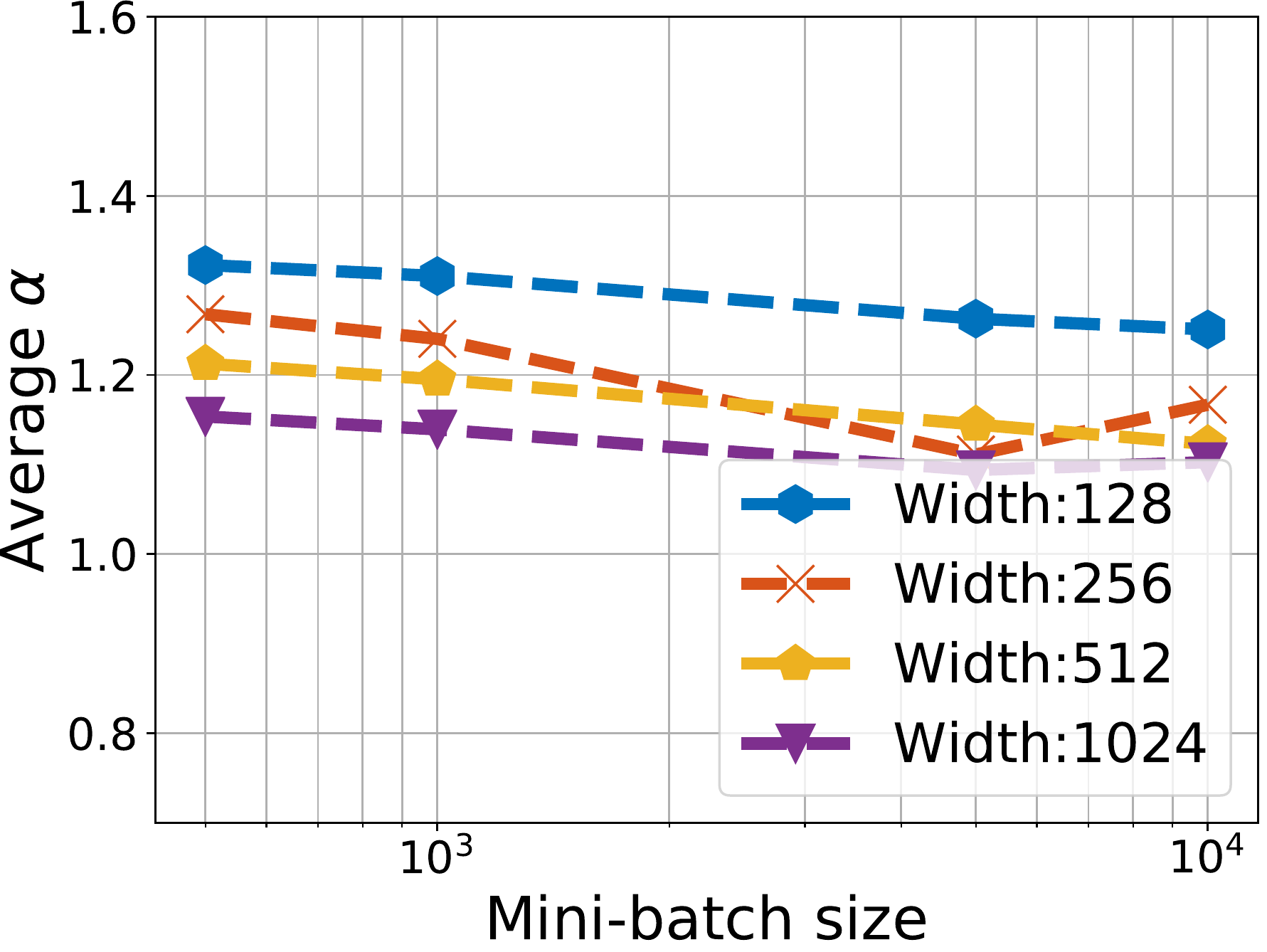}}
    \vspace{-10pt}
    \caption{Estimation of $\alpha$ for varying minibatch size.}
    \label{fig:exp_fc_mbscale}
\end{figure}

\textbf{Effect of the minibatch size: }
In our second set of experiments, we investigate the effect of the size of the minibatch on $\alpha$. We focus on the FCN and monitor the behavior of $\alpha$ for different network and minibatch sizes $b$. Figure~\ref{fig:exp_fc_mbscale} illustrates the results. These rather remarkable results show that, as opposed to the common belief that the gradient noise behaves similar to a Gaussian for large $b$, the tail-index does not increase at all with the increasing $b$. We observe that $\alpha$ stays almost the same when the depth is $2$ and it moves in a small interval when the depth is set to $4$. We note that we obtained the same the train and test accuracies for different minibatch sizes.


\begin{figure}[t]
    \centering
    \vspace{5pt}
    \subfigure[MNIST]{
    \includegraphics[width=0.33\columnwidth]{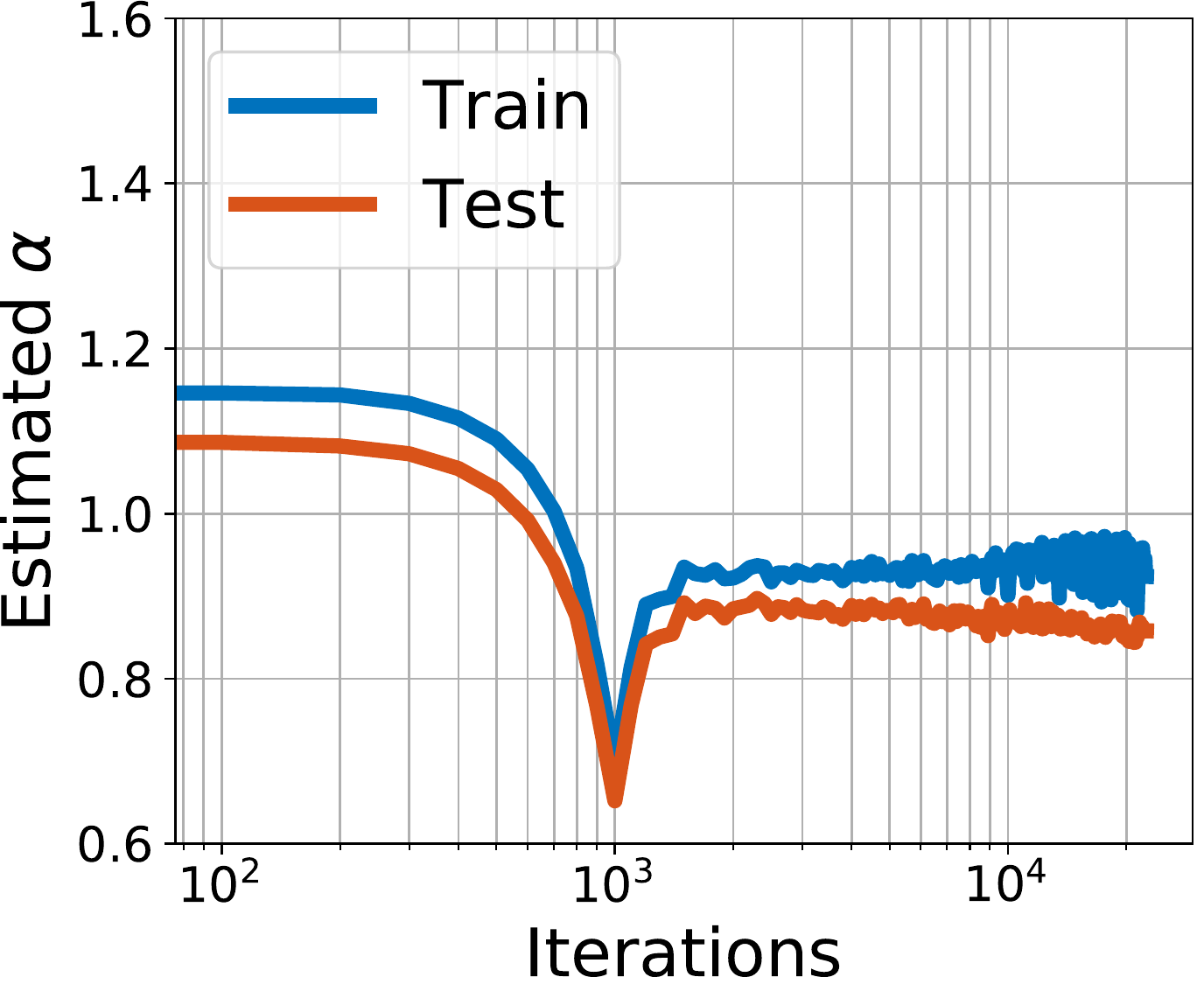}
    \includegraphics[width=0.33\columnwidth]{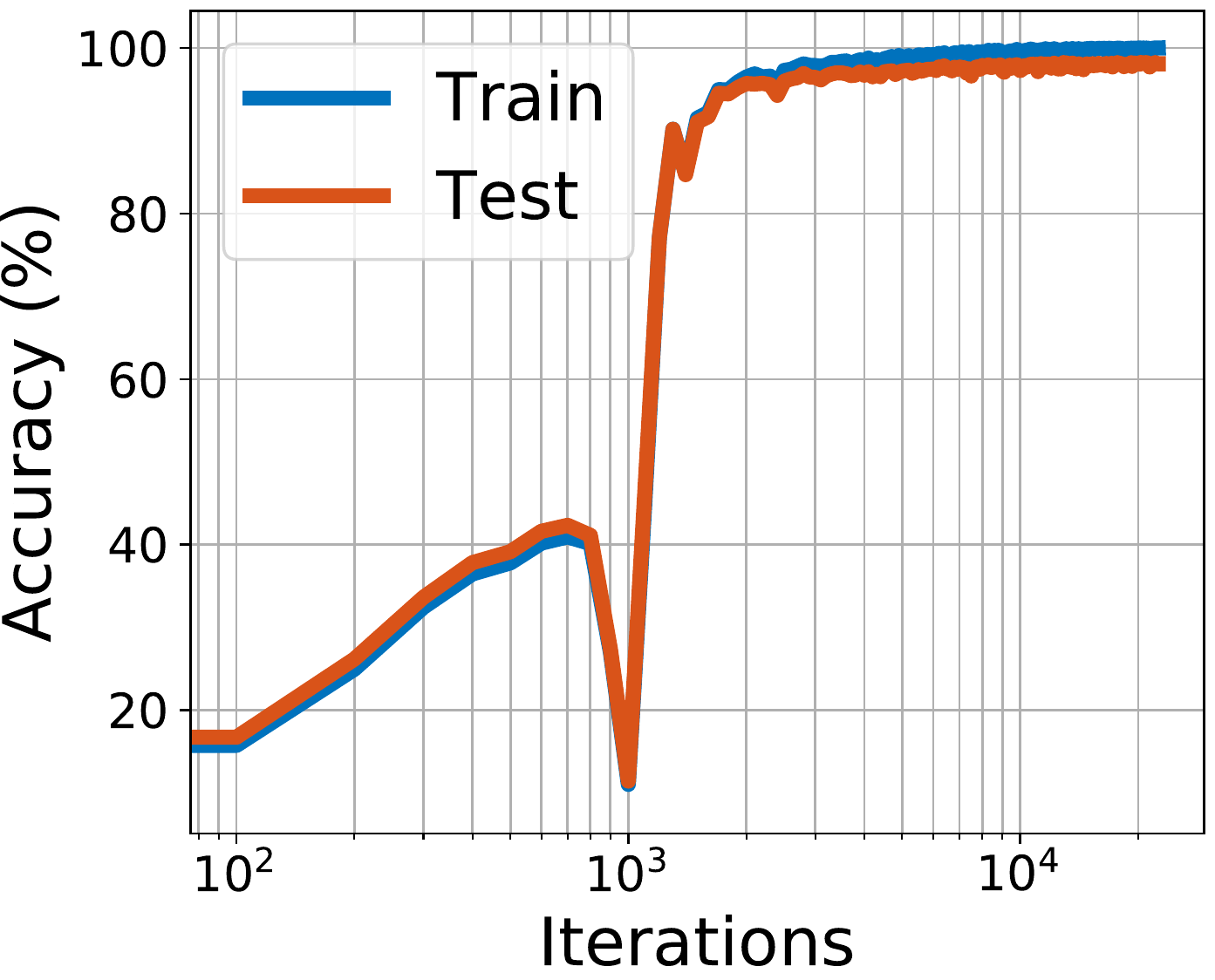}
    \includegraphics[width=0.33\columnwidth]{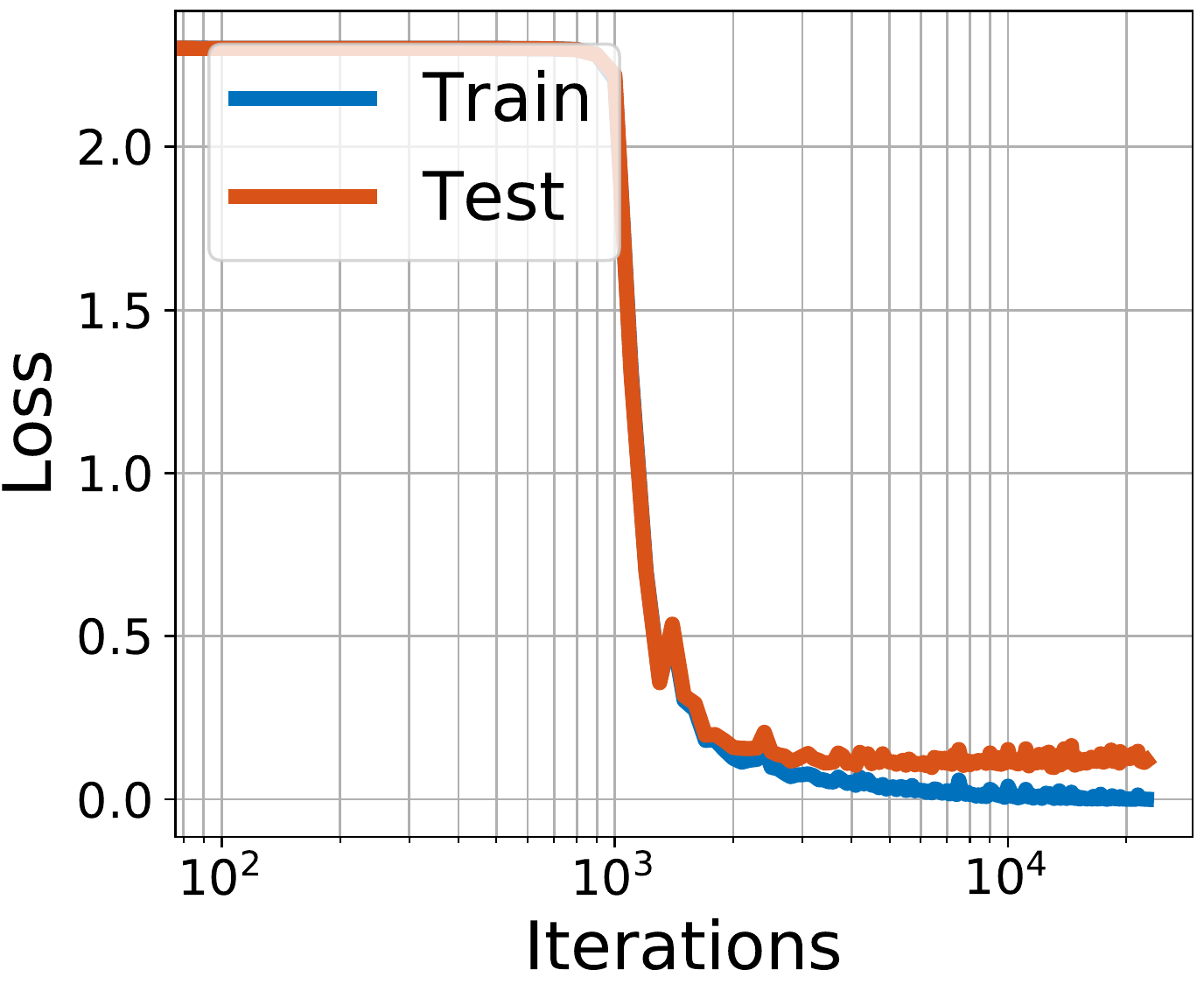}
    }
    \subfigure[CIFAR10]{
    \includegraphics[width=0.33\columnwidth]{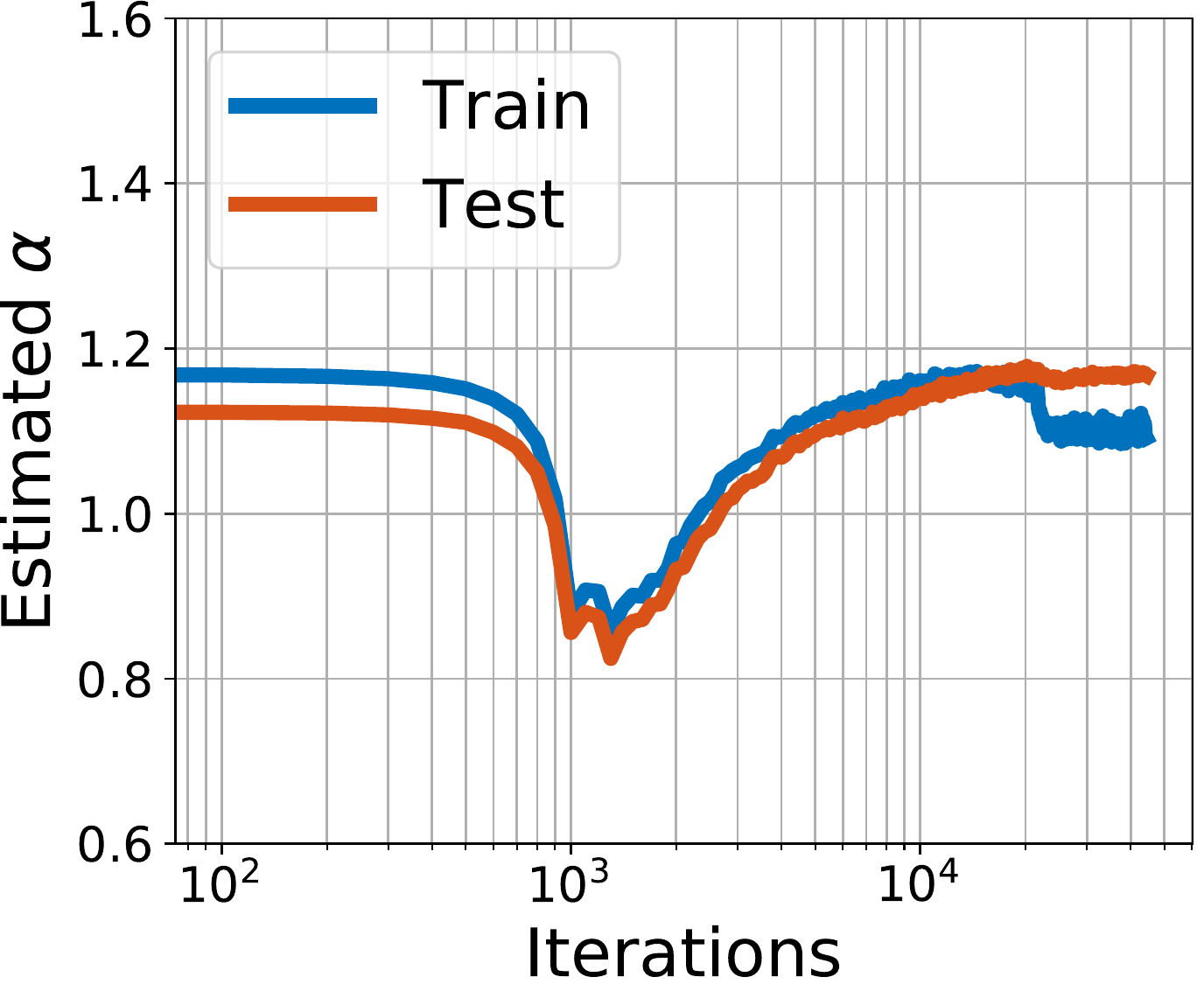}
    \includegraphics[width=0.33\columnwidth]{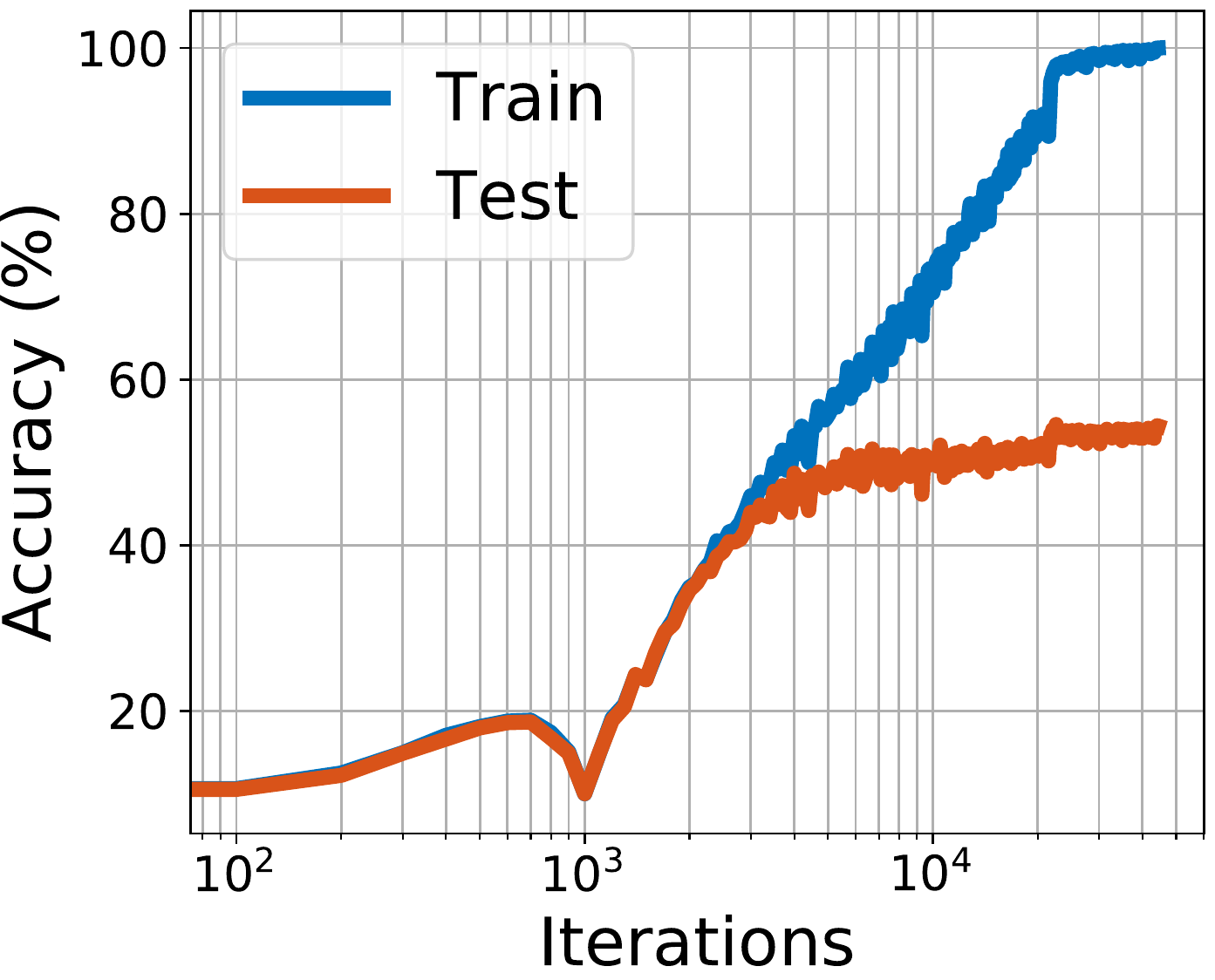}
    \includegraphics[width=0.32\columnwidth]{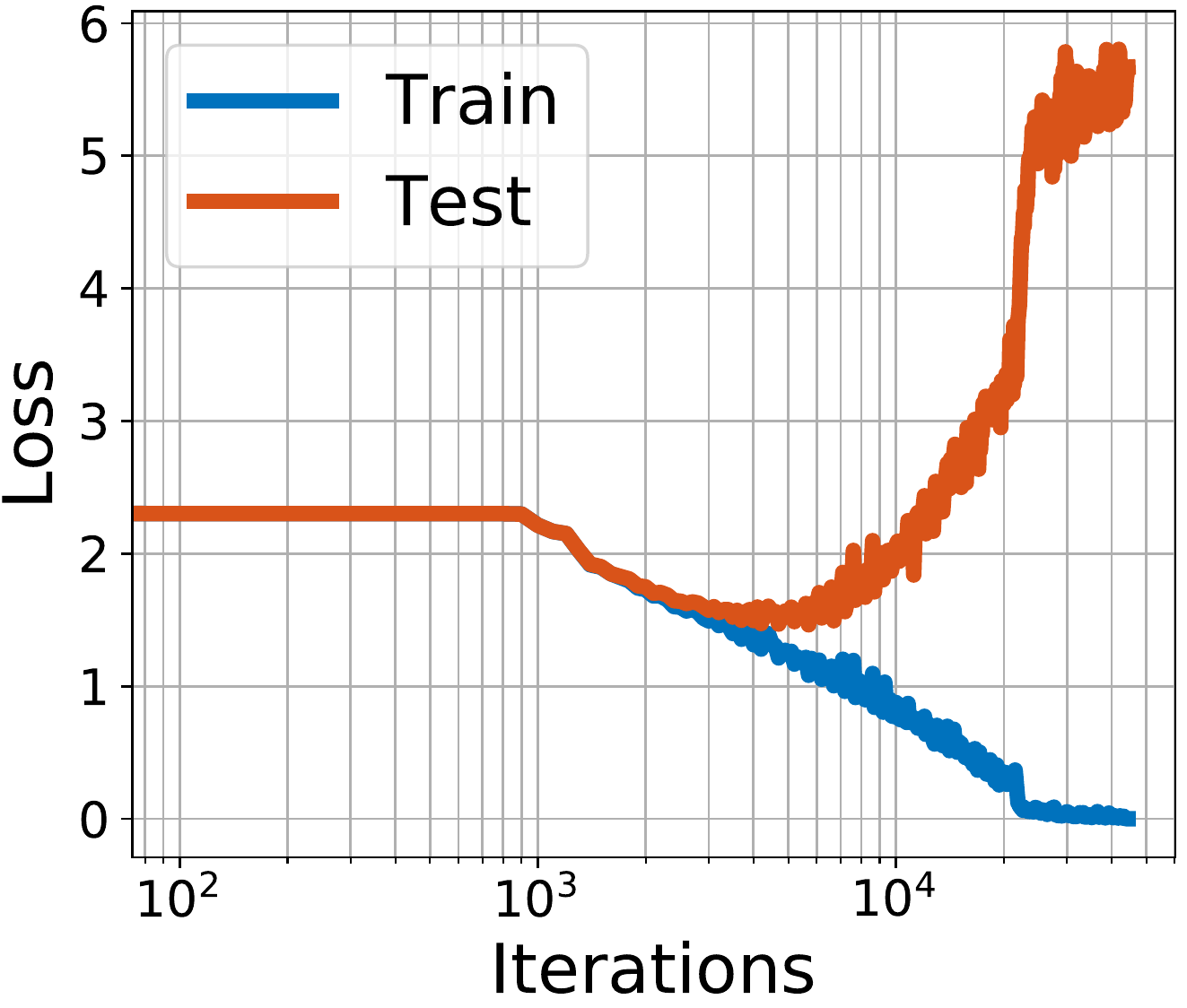}
    }
    \vspace{-12pt}
    \caption{The iteration-wise behavior of of $\alpha$ for the FCN.}
    \vspace{-5pt}
    \label{fig:exp_iter_fc}
\end{figure}

\textbf{Tail behavior throughout iterations: }
So far, we have focused on the last iterations of SGD, where $\alpha$ is in a stationary regime. In our last set of experiments, we shift our focus on the first iterations and report an interesting behavior that we observed in almost all our experiments. As a representative, in Figure~\ref{fig:exp_iter_fc}, we show the temporal evolution of SGD for the FCN with $9$ layers and $512$ neurons/layer. 

The results clearly show that there are two distinct phases of SGD (in this configuration before and after iteration $1000$). In the first phase, the loss decreases very slowly, the accuracy slightly increases, and more interestingly $\alpha$ rapidly decreases. When $\alpha$ reaches its lowest level, the process possesses a jump, which causes a sudden decrease in the accuracy. After this point the process recovers again and we see a stationary behavior in $\alpha$ and an increasing behavior in the accuracy.  

The fact that the process has a jump when $\alpha$ is at its smallest value provides a strong support to our assumptions and the metastability theory that we discussed in the previous section. Furthermore, these results further strengthen the view that SGD crosses barriers at the very initial phase.  On the other hand, our current analysis is not able to determine whether the process jumps in a different basin or a `better' part of the same basin and we leave it as a future work. 

%% file: conclusion.tex
\section{Conclusion and Open Problems}
\label{sec:conc}


We investigated the tail behavior of the gradient noise in deep neural networks and empirically showed that the gradient noise is highly non-Gaussian. This outcome enabled us to analyze SGD as an SDE driven by a L\'{e}vy motion and establish a bridge between SGD and existing theoretical results, which provides more illumination on the behavior of SGD, especially in terms of choosing wide minima.

This study also brings up interesting open questions and future directions: (i) While the current metastability theory applies for the continuous-time processes, the behavior of the discretized process and its dependence on the algorithm parameters (e.g., the step-size, minibatch size) are not clear and yet to be investigated. (ii) We observe that, especially during the first iterations, the tail-index depends on the current state $\wb_k$, which suggests analyzing SGD as a stable-like process \cite{bass1988uniqueness} where the tail-index can depend on time. However, the metastability behavior of these processes are not clear at the moment and its theory is still in an early phase \cite{kuhwald2016bistable}. (iii) Furthermore, an extension of the current metastability theory that includes minima with zero modes is also missing and appears to be challenging yet crucial direction of future research. 





%% file: alpha_estim.tex
\section{Measuring the Accuracy of the Tail Estimator}

\begin{figure}[h]
    \centering
    \includegraphics[width=0.4\columnwidth]{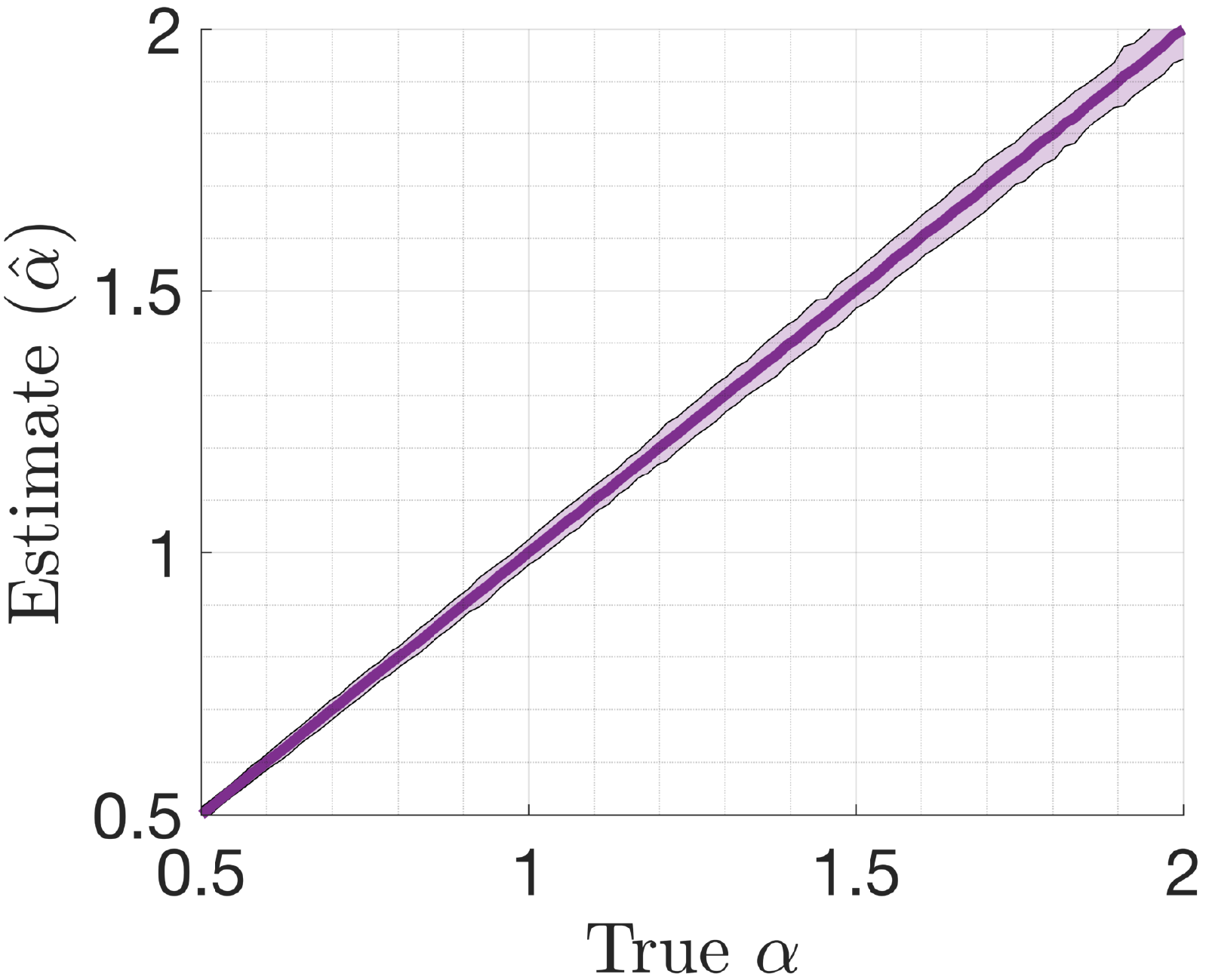} 
    \caption{Estimation of $\alpha$. 
    }
    \label{fig:exp_synth}
\end{figure}

In order to verify the accuracy of this estimator, we conduct a preliminary experiment, where we first generate $K = k \times l$ many $\sas(1)$ distributed random variables with $k= 100$, $l =1000$ for $100$ different values of $\alpha$. Then, we estimate $\alpha$ by using $\hat{\alpha} \triangleq (\widehat{\phantom{a}\frac1{\alpha}\phantom{a}})^{-1}$. We repeat this experiment $100$ times for each $\alpha$. As shown in Figure~\ref{fig:exp_synth}, the estimator is very accurate for a large range of $\alpha$. Due to its nice theoretical properties and empirical validity, we choose this estimator in our experiments. 

%% file: width_flat.tex

\section{Making sense of different basins on flat landscapes}
\label{sec:width_flat}

In this section we attempt to connect the basin hopping perspective on flat landscapes

\begin{itemize}
    \item toy experiment to show an overparametrized flat landscape
    \item decomposition of the weight space into two parts
    \item cite redundant op cohen, dlp, criteo guys etc...n
\end{itemize}

Take $f(w) = w^2$ and $\hat f(w_1, w_2)=(w_1w_2)^2$ plot the graph of $f$ and plot the level curves of $\hat f $. take random initial points and run GD, then run GD + noise whose alpha should range from 2 to 0.5. repeat this experiment with many different initial points and collect statistics. Here the width can be measured by the distance from the zero point. 

at a solution where wolog $w_1=0$, grad is zero, one eigenvalue of the hessian is zero and the second eigenvalue is $2w_2^2$ which can be the proxy for `width'

test to see if the one dimensional theory fits into this framework where, in some sense, there is a continuum many minima, each of which with different \textit{width}.

The problem is convex but various line segments are non-convex. This strurcture is similar to what happens in deep learning. In DL we have a convex loss function (MSE NLL hinge etc...) and a non-linear output function. It doesn't have to be super nonconvex in the strict sense, the landscape is skewwed and ill conditioned in many ways but it can essentially convex once one looks at it from the global weight space point of view. However, navigating this ill-conditioned landscape is similar to navigating different basins on a landscape that is formed of many isolated minima.

%% file: discretization.tex

\section{Metastability in the discretized regime}